\documentclass{article}
\usepackage{ijcai24}
\usepackage{colortbl}
\usepackage{times}
\usepackage{soul}
\usepackage{url}
\usepackage[hidelinks]{hyperref}
\usepackage[utf8]{inputenc}
\usepackage[small]{caption}
\usepackage{graphicx}
\usepackage{amsmath}
\usepackage{amsthm}
\usepackage{booktabs}
\usepackage{algorithm}
\usepackage{algorithmic}
\usepackage[switch]{lineno}

\usepackage{subfig}
\usepackage{multirow}
\usepackage{enumitem}
\usepackage{makecell}
\usepackage{threeparttable}
\usepackage{amsfonts,mathtools}
\usepackage{mathrsfs}
\usepackage{float}

\newcommand{\M}{GPFN}

\newcommand{\redfont}[1]{{\textcolor{red}{#1}}}

\title{Infinite-Horizon Graph Filters: Leveraging Power Series \\ to Enhance Sparse Information Aggregation }

\author{
Ruizhe Zhang$^1$\textsuperscript{*}\and
Xinke Jiang$^1$\textsuperscript{*}\and
Yuchen Fang$^2$\footnote{Ruize Zhang, Xinke Jiang, and Yuchen Fang contributed equally to this research.} \and
Jiayuan Luo$^2$\and
Yongxin Xu$^1$\and \\
Yichen Zhu$^3$\and 
Xu Chu$^1$\textsuperscript{\textdaggerdbl}\and
Junfeng Zhao$^1$\textsuperscript{\textdaggerdbl}\footnote{Junfeng Zhao is also at the Big Data Technology Research Center, Nanhu Laboratory, 314002, Jiaxing.}\And
Yasha Wang$^1$\footnote{Correspounding Authors.}\\
\affiliations
$^1$Key Laboratory of High Confidence Software Technologies (Peking University) \\ Ministry of Education; School of Computer Science, Peking University\\
$^2$No affaliation\\
$^3$University of Toronto\\
\emails
\{nostradamus, xinkejiang\}@stu.pku.edu.cn,
fyclmiss@gmail.com,
joyingluo@foxmail.com,
xuyx@stu.pku.edu.cn,
yichen\_zhu@foxmail.com,
chu\_xu@pku.edu.cn,
zhaojf@pku.edu.cn,
Wangyasha@pku.edu.cn,
}
\begin{document}

\maketitle

\begin{abstract}
Graph Neural Networks (GNNs) have shown considerable effectiveness in a variety of graph learning tasks, particularly those based on the message-passing approach in recent years. However, their performance is often constrained by a limited receptive field, a challenge that becomes more acute in the presence of sparse graphs. In light of the power series, which possesses infinite expansion capabilities, we propose a novel \underline{G}raph \underline{P}ower \underline{F}ilter \underline{N}eural Network (GPFN) that enhances node classification by employing a power series graph filter to augment the receptive field. Concretely, our GPFN designs a new way to build a graph filter with an infinite receptive field based on the convergence power series, which can be analyzed in the spectral and spatial domains. Besides, we theoretically prove that our GPFN is a general framework that can integrate any power series and capture long-range dependencies. Finally, experimental results on three datasets demonstrate the superiority of our GPFN over state-of-the-art baselines\footnote{Code is anonymously available at https://github.com/GPFN-Anonymous/GPFN.git}.

\end{abstract}



\section{Introduction}
\label{introduction}

Graph neural networks (GNNs) have attracted significant attention in the research community due to their exceptional performance in a variety of graph learning applications, including social analysis~\cite{qin2022next,sn_IJCAL} and traffic forecasting~\cite{SPGCL,gao2023spatiotemporal,sttd,fang2023spatio}. A prevalent method involves the use of  message-passing~\cite{GCN,hamilton2017inductive} technique to manage node features and the topology of the graph. Various layer types~\cite{gin,chebnet} like graph convolutional (GCN) ~\cite{GCN} and graph attention layers (GAT)~\cite{GAT} enable GNNs to capture complex relationships, enhancing their performance across multiple domains. However, despite their advancements in graph representation learning, message-passing-based GNNs still face certain limitations.

\begin{figure}[ht]
  \centering
  \includegraphics[scale=0.15]{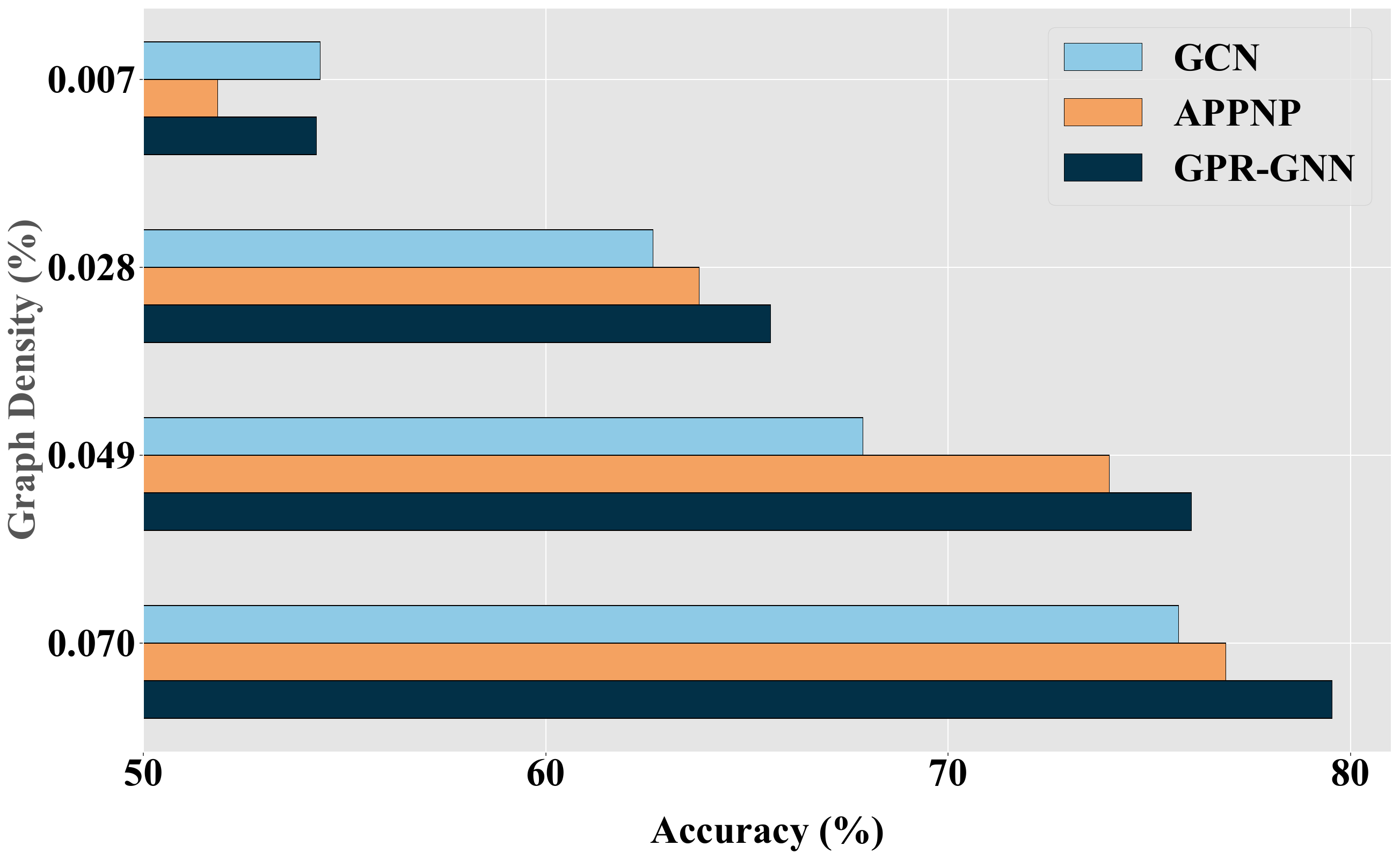}
  \vspace{-5pt}
  \caption{The influence of GCN~\protect\cite{GCN}, APPNP~\protect\cite{APPNP}, and GPR-GNN~\protect\cite{GPRGNN} on the node classification task under different sparse situations.}
  \label{fig:inrto_decline.png}
  \vspace{-10pt}
\end{figure}

\textbf{i) Long-range Dependencies:} Balancing the trade-off between the receptive field size and feature distinctiveness is a challenging aspect in GNNs. On the one hand, deeper GNNs~\cite{GPRGNN,ResGCN,BernNET} offer a larger receptive field, which allows for the incorporation of information from a broader range of the graph. However, this comes with the downside of feature homogenization across nodes, leading to the phenomenon known as over-smoothing. On the other hand, GNNs with shallower structures~\cite{rusch2023survey}, while avoiding over-smoothing, face limitations in capturing long-range dependencies due to their smaller receptive field. This limitation is particularly significant in real-world graphs, such as social networks or protein interaction networks, where understanding distant node relationships is crucial. 

\textbf{ii) Graph Sparsity:} A sparse graph is a type of graph in which the number of edges is relatively low compared to the total number of possible edges. Such graphs are common in real-world scenarios. For instance, \cite{power_lay_2,power_lawer_1} reveals that the degree distribution of real-world graphs typically follows a power-law distribution. In particular, the citation network-based Cora and Citeseer datasets exhibit remarkable levels of sparsity, measured at 99.93\% and 99.96\%, respectively. Because sparse graphs contain less explicit information due to fewer edges, it is hard to mine effective representations, even for contemporary GNNs. As shown in Figure~\ref{fig:inrto_decline.png}, as the number of edges decreases gradually, the performance of GNNs drops correspondingly, indicating that learning effective representations on sparse graphs remains an unresolved challenge.

In this paper, we endeavor to address the challenges previously mentioned by explicitly modeling dependencies over an infinite range within a single layer. This strategy effectively harnesses valuable information in sparse graphs and captures long-range dependencies without incurring the over-smoothing issue. Specifically, inspired by the impressive capability of power series for infinite expansion, we propose a novel method named \textbf{\underline{G}}raph \textbf{\underline{P}}ower \textbf{\underline{F}}ilter \textbf{\underline{N}}eural Network (GPFN). A noteworthy point is that employing a standard power series graph filter can lead to a substantial computational burden. Existing approaches, such as BernNet~\cite{BernNET}, opt to truncate the K-order polynomial to simplify the complexity, but these methods might lose long-range information. In contrast, GPFN is designed to construct the graph filter using convergent power series from the spectral domain. This approach ensures the preservation of the infinite modeling capability of power series without information loss. We substantiate the effectiveness of our proposed GPFN with both theoretical and empirical evidence.

In summary, our contributions are listed as follows:

\begin{itemize} [leftmargin=*]
    \item Our proposed framework GPFN utilizes convergent infinite power series derived from the spectral domain for aggregating long-range information, which significantly mitigates the adverse impacts associated with graph sparsity.
    \item We analyze GPFN from a spatial domain perspective, focusing on its ability to effectively capture long-range dependencies. Additionally, we provide theoretical evidence demonstrating that our GPFN is not only capable of achieving exceptional performance using shallow layers but also effectively integrates various power series.
    \item We validate our GPFN through experiments on three real-world graph datasets, tested across various sparse graph settings. The experimental results demonstrate the advantages of our GPFN over state-of-the-art baselines, especially in contexts of extreme graph sparsity.
\end{itemize}

    

\section{Related Work}
\label{related work}



\paragraph{Monomial graph filters.} These filters are mainly aimed at filtering information between two layers, without introducing more layer parameters. 
Spectral GNNs are grounded in the concept of the graph Fourier filter, as introduced by~\cite{gsp,gcn_raw}, wherein the eigenbasis of the graph Laplacian is analogously employed.  
Subsequently, GCN~\cite{GCN} substitutes the convolutional core with first-order Chebyshev approximation. Other monomeric graph filters are GAT~\cite{GAT},  GIN~\cite{gin}, AGE~\cite{AGE} and SGC~\cite{SGC}.

\paragraph{Polynomial graph filters.} 
Polynomial filters encompass ResGCN~\cite{ResGCN} (including those GNNs employing long-range residual connection such as Graph Transformer~\cite{NEURIPS2021_6e67691b}), BernNet~\cite{BernNET}, and GPR-GNN~\cite{GPRGNN}, etc.. ResGCN employs residual connections to facilitate information transfer between different layers, alleviating the over-smoothing issues. However, ResGCN focuses more on intermediate information, neglecting the significance of proximal information. BernNet utilizes Bernstein polynomials to aggregate information across different layers. Nevertheless, due to layer constraints, BernNet struggles to extend its reach to more distant information, and parameter learning becomes more challenging. APPNP~\cite{APPNP} and GRAND~\cite{2022GRAND} utilize feature propagation but within a limited number of hops. GPR-GNN unifies the representation of parameters between different layers in a general polynomial formula, therefore APPNP, ResGCN and BernNet can all be regarded as special case of GPR-GNN. However, GPR-GNN requires learning relative control parameters between layers. Similar to BernNet, the receptive field of view is limited, and parameter learning poses a significant challenge.


\section{Preliminaries}
\label{prelinminaries}

\subsection{Problem Formulation}
\paragraph{\textit{Definition 1. (Sparse graph)}}
Given a graph $\mathcal{G} = (\mathcal{V, E, Y}, A, X)$, $\mathcal{V}=\{v_1, \cdots, v_N \}$ is the set of nodes, and $\mathcal{E}\subseteq \mathcal{V} \times \mathcal{V}$ is the edge set. $\mathcal{Y}$ is the set of labels for node in $\mathcal{V}$. 
$A\in \mathbb{R}^{N\times N}$ denotes the adjacency matrix, where $A_{ij}>0$ if $(v_i, v_j)\in \mathcal{E}$ and $A_{ij}=0$ if $(v_i, v_j) \notin \mathcal{E}$. $X\in \mathbb{R}^{N\times D}$ is the attribute matrix, and $D$ is the number of attribute dimensions. 
If there exists $|\mathcal{E}|\ll|\mathcal{V}^2|$, we call $\mathcal{G}$ a sparse graph.

\paragraph{\textit{Definition 2. (Label prediction for sparse graph)}}
Given a sparse graph $\mathcal{G} = (\mathcal{V, E, Y}, A, X)$, and the node set is divided into a train set and a test set, i.e., $\mathcal{V} = \mathcal{V}_{\rm{train}} \cup \mathcal{V}_{\rm{test}}$. 
The label $y_i$ of node $v_i$ can be observed only if $v_i \in \mathcal{V}_{\rm{train}}$. The goal of label prediction is to predict test labels $\mathcal{Y}_{\rm{test}} = \{y_i|v_i \in \mathcal{V}_{\rm{test}} \}$. We utilize a two-layer GNN for downstream tasks and predict the label $\hat Y$ as $\hat{Y} = \text{GNN}(X, A)$. And the surprised Cross-Entropy Loss function is:$ \quad
\min_{\rho} L_{pre} =
\frac{1}{N} \sum_i \sum_{c=1}^{|Y|} -y_{ic}\log(\hat{y}_{ic}),
$
here $|Y|$ is the number of classes, and $y_{ic}$ is 1 if node $v_i$ belongs to class $c$, else it is 0. $\hat{y}_{ic}$ is the predicted probability that node $v_i$ belongs to class $c$. $\rho$ is the parameter of the GNN predictor.


\subsection{Revisiting Graph Neural Networks}
We revisiting GNNs from two perspectives: 
\paragraph{\textbf{\textit{(Spatial Domain)}}} The message passing-based GCN~\cite{GCN} can be formed as follows:
\begin{equation}
  \footnotesize
    H^{(0)} = X, \quad H^{(l+1)} = \sigma(\hat A H^{(l)}W^{(l)}),
    \label{eq:MPNN-aggretation}
\end{equation}
where $\hat A$ is the aggregation matrix. Particularly, $\tilde A_{sym} = \tilde{D}^{-\frac{1}{2}}\tilde{A}\tilde{D}^{-\frac{1}{2}}$, where $\tilde{A} = A\,+\,I$ and $\tilde{D}$ are the adjacency matrix with the identity matrix-based self-loop and the degree matrix of $\tilde{A}$. $H^{(l)} \in \mathbb{R}^{N\times D}$ is the node representation matrix at $l$-th layer and $\sigma(\cdot)$ denotes an activation function.
\paragraph{\textbf{\textit{(Spectral Domain)}}} 
The spectral convolution~\cite{Shuman_2013} of attribute $X$ and filter $F_{\gamma}$ can be formulated as: 
\begin{equation}
  \footnotesize
  F_{\gamma} * X = U\bigl( (U^T F_{\gamma})\odot(U^T X)\bigl) = U F_{\gamma}(\Lambda) U^T X
    ,
    \label{eq:spectral convolution}
\end{equation}
where $*$ is graph convolution and $\odot$ is Hadamard product. Note that the decomposition of aggregation matrix$ \hat{A} = U \Lambda U^T$ can be used to obtain eigenvectors as Fourier bases and eigenvalues as frequencies.


\subsection{Eigenvalue of Aggregation Matrix}
Previous studies \cite{AGE} reveal that the Rayleigh Quotient can be used to calculate the lower bound and upper bound of eigenvalues of $\hat{A}$, that is $\lambda_{min}=\min(R)\leq \max(R) = \lambda_{max}$~\cite{AGE}. Thus, for node $v_i$, its Rayleigh Quotient is $R(\hat{A}, u_i) = \lambda_i$, and we let $q_i=\widetilde{D}^{-1/2}u_i$, where $U=\text{diag}(u_1,\ldots,u_n)$ is eigenvalue decomposition of $\hat{A}$. Because $R(\widetilde{L}_{sym}, u_i)$ is a division of two quadratic forms, the eigenvalues are non-negative. We prove that the maximum of $\widetilde{L}_{sym}$ eigenvalue is $2$ iff the graph is bipartite as shown in Appendix A. Therefore, for $ \widetilde{L}_{sym} = I+\tilde{A}_{sym}$, the eigenvalues of the $\tilde{A}_{sym}$ satisfy the following equation: $1 = \lambda_1 \geq \lambda_2 \geq \cdots \geq \lambda_N > -1$, as previous study reveals~\cite{spectral}.

\section{Methodology}
\label{method}
\begin{figure}[htbp]
  \centering
  \includegraphics[width=1.0\linewidth]{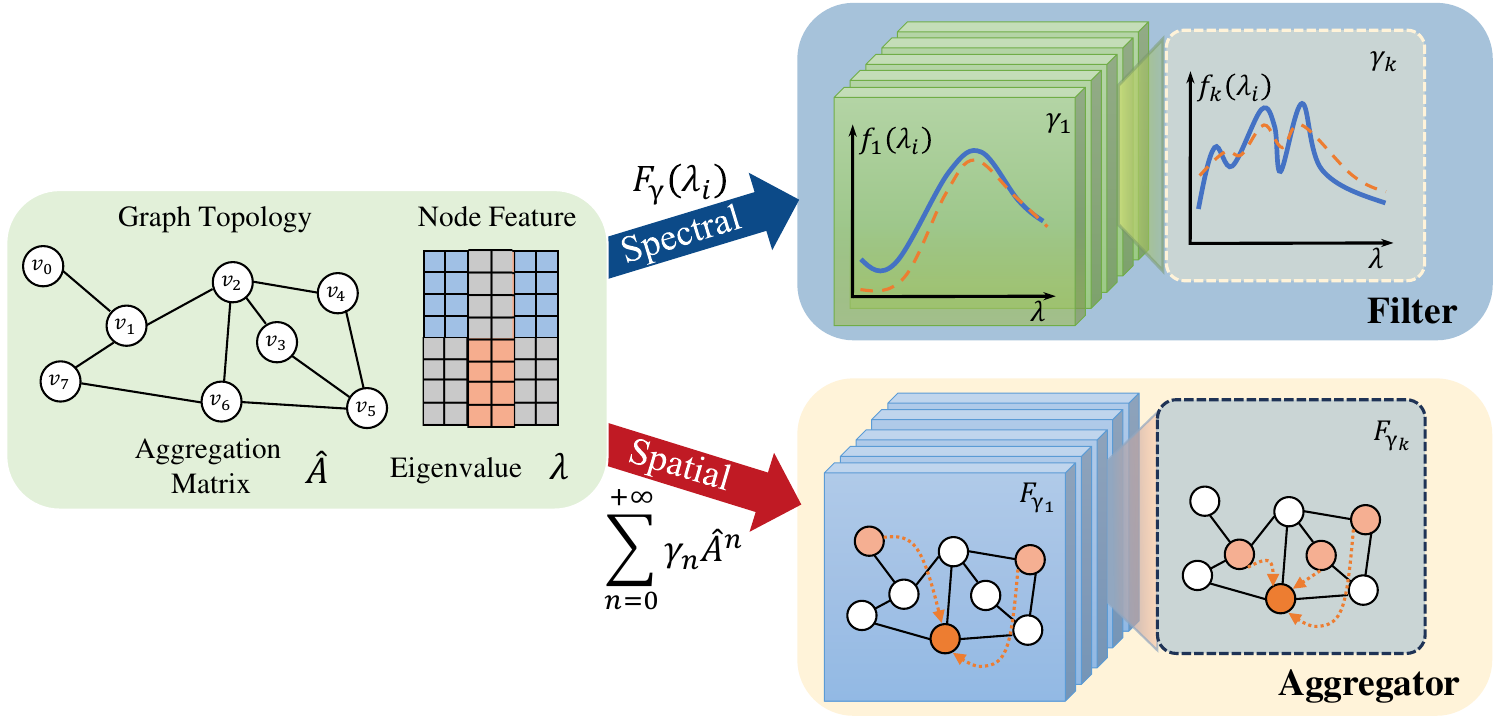}
  \caption{The overall framework of \M. For arbitrary power series over the aggregation matrix, we can design corresponding filters in the spectral domain, which also serve as an infinite aggregator in the spatial domain.
}
  \label{fig:main_model.png}
\vspace{-0.3cm}
\end{figure}

In this section, we detail our proposed GPFN. Initially, in Section \ref{Generalized Filter}, we elucidate the methodology for designing a graph filter based on a power series. Subsequently, in Sections \ref{Effectiveness Analysis} and \ref{Discussion of Filter Type}, we introduce three fundamental power series filters and substantiate their effectiveness as well as the rationale behind the choice of hyperparameters $\beta_0$. Ultimately, we juxtapose and analyze the relations between our Power Series Filters and the preceding research. In general, there are two perspectives as shown in Figure~\ref{fig:main_model.png} – \textbf{spectral} and \textbf{spatial} domains to support the analysis of Power Series Filters:
\begin{itemize}[leftmargin=*]
\item \emph{\textbf{(Spectral Domain)} A flexible graph filter framework}.
We demonstrate that a variety of filter types, such as low-pass, high-pass, or band-pass, can be conveniently devised by simply adjusting the filter coefficients $\gamma_n$ (refer to Section \ref{Generalized Filter}). Furthermore, we exhibit that other polynomial filters can be induced into our framework, thereby affirming its extensive applicability and adaptability.
\item \emph{\textbf{(Spatial Domain)} An infinite information aggregator}.
Filters constructed via power series possess the capability to aggregate neighborhood information across an infinite number of hops with variant weights, thereby enlarging the graph’s receptive field.
\end{itemize}

\vspace{-10pt}

\subsection{Generalized Filter}
\label{Generalized Filter}
\subsubsection{Foundation of Power Series}
To grasp the fundamentals of \M, it is essential to review the concept of power series. In mathematics, a power series, exemplified here as a single variable, is an infinite series of the standard form:
\begin{equation}
\footnotesize
\sum_{n=0}^{+\infty} a_n x^n =a_0 + a_1 x + \cdots + a_i x^i + \cdots,  \label{eq:ps_standard_definition}
\end{equation}
where $i\in \mathbb{N}^+$, $a_n$ represents the coefficient of the $n$-th term. 
 When this power series converges and variable $x$ is constrained within the convergence region $[-r,r]$ with boundary $ r \in \mathbb{R^+}$, this power series approaches a finite limit. In this case, it can be regarded as the expansion of an infinitely differentiable function $f(x) = \sum_{n=0}^{+\infty} a_n x^n, x \in [-r,r]$. The key to our work is to design a reasonable function $f(x)$. To achieve this, we leverage the power series to perform graph convolution, as illustrated in the next subsection.

\subsubsection{Power Series Graph Filters}
In \M, We reparameterize the variables $x$ as $x=\beta_0 r \hat{A}$, and $a_n$ to $a_n=\frac{\gamma_n}{r^n \beta_0^n}$, where $r$ is imported to restrict the variable within the convergence region, $\beta_0 \in (0,1)$ is blend factor to control the strength of filters, and $\gamma_n$ representing different weights of each hop is the filter coefficient for designing different kinds of graph filters. Then we derive the generalized formula for the power series filter:
\begin{equation}
\footnotesize
F_\gamma(\hat{A}) = \sum_{n=0}^{+\infty} \gamma_n \hat{A}^n, 
    \label{eq:ps-filter definition}
\end{equation}
where $F_\gamma(\hat{A})$ represents the general term of power series filter. And $\hat{A}^n$ captures long-range dependencies between nodes within $n$-hops~\cite{SGC}. The convergence region in Eq.~(\ref{eq:ps-filter definition}) is rescale to $(-\frac{1}{\beta_0},\frac{1}{\beta_0})$, thus we have to choose aggregation matrix $\hat{A}$ with eigenvalues fall in this region.

Let $\hat{A}=U \Lambda U^T$ be the eigenvalue decomposition of $\hat{A}$, where $U\in \mathbb{R}^{N\times N}$ consists of eigenvectors and $\Lambda=\text{diag}(\lambda_1, \lambda_2, \cdots, \lambda_n)$ is a diagonal matrix of eigenvalues. Because matrix $U$ is a orthogonal matrix, $U U^T= U^T U=I$, we have $\hat{A}^n=(U \Lambda U^T)^n = U \Lambda^n U^T$. After applying the power series element-wise, we have:
\begin{equation}
 \footnotesize
 \begin{aligned}
 \footnotesize
&F_\gamma(\hat{A}) = \sum_{n=0}^{+\infty} \gamma_n \hat{A}^n = U (\sum_{n=0}^{+\infty} \gamma_n \Lambda^n)U^T 
\\ &= U \text{diag}(\sum_{n=0}^{+\infty} \gamma_n \lambda_1^n,\sum_{n=0}^{+\infty} \gamma_n \lambda_2^n, \ldots ,\sum_{n=0}^{+\infty} \gamma_n \lambda_n^n)U^T
=U F_\gamma(\Lambda)U^T.
\end{aligned}
    \label{eq:ps-filter computation}
    \nonumber
\end{equation}

Therefore, by selecting different power series bases, we can design different forms of graph filters, as illustrated in Table~\ref{tab:filters}. Commonly, we employ adjacency matrix variants of $\tilde{A}_{sym}$ or $\tilde{L}_{sym}$ as the aggregation matrix, renamed as $\hat{A}$. In this way, we propose an efficient way to build a graph filter based on any convergence power series $F_\gamma(\cdot)$ from the spectral domain. 
Meanwhile, by expanding $F_\gamma(\cdot)$, we observe that \M~ assign different weights to different hops of neighbors until infinite, allowing each node to collect and integrate information from its more distant neighbors. Thus, \M~ is also equivalent to an infinite information aggregator from the spatial domain. Besides, if we assume that as $K\to+\infty$, $H^{(K)}$ would be homogeneous to boundary $\mathcal{B}+o_K(1)$, and we find that $\lim_{K\to+\infty}|{(H^{(K)}-\mathbf{1}\mathcal{B}})| \propto \lim_{K\to+\infty}|{(UF_{\gamma}(\Lambda)^{K}U^T-\mathbf{1}\mathcal{B}})|=0 $ is indeed a low order infinitesimal to other graph filters such as GCN $\lim_{K\to+\infty}|{(U\Lambda^{K}U^T-\mathbf{1}\mathcal{B}})|=0$ as well as GPR-GNN's. That's to say, GPFN has a slower convergence speed than other baselines with $K$ increases, which can effectively alleviate the problem of over-smoothing and could construct GNNs deeper, as we demonstrate in Appendix B. 

\begin{table}[htbp]
    \aboverulesep=0ex
    \belowrulesep=0ex
	\centering
	\footnotesize
   \setlength\tabcolsep{1.1pt}   
   \renewcommand\arraystretch{1.5}
	\resizebox{1.0\linewidth}{!}{\begin{tabular}{l|l|c c c c c}
		\toprule
		&\textbf{Filter name} & \textbf{Filter type} &  {\textbf{$F_\gamma(\hat{A})$}} & ${\textbf{\( \gamma_n \) }}$ & \textbf{$\hat{A}$} & \textbf{Receptive Field}  \\
        \midrule
	\multirow{3}{*}{\rotatebox[origin=c]{90}{Monomial}} 
        &\texttt{GCN} & low-pass & $(I-\hat{A})^{K}$ & $ / $ & ${\tilde{A}_{sym}}$ & $K$\\
        &\texttt{AGE} & low-pass & $\alpha \hat{A}$ & $/$ & ${\tilde{L}_{sym}}$ & $1$\\
        &\texttt{SGC} & low-pass & $\hat{A}^K$ & $ / $ & ${\tilde{A}_{sym}}$ & $K$\\
        \midrule
        \multirow{5}{*}{\rotatebox[origin=c]{90}{Polynomial}} 
        &\texttt{APPNP} & low-pass & $\alpha[I-(1-\alpha)\hat{A}]^{-1}$ & $\alpha (1-\alpha)^n$ & ${\tilde{A}_{sym}}$ & $K$\\
        &\texttt{ChebNet} & low-pass & $/$ & $\gamma_k \cos(k \arccos(1-\lambda)$ & ${\tilde{L}_{sym}}$ & $K$\\
        &\texttt{Res-GCN} & low-pass & $/$ & $\binom{K}{n}$ & ${\tilde{A}_{sym}}$ & $K$\\
        &\texttt{GPR-GNN} & comb-pass & $/$ & $\gamma_k (1-\lambda)^k$ & ${\tilde{A}_{sym}}$ & $K$\\
        &\texttt{HiGCN} & comb-pass & $/$ & $\gamma_{p,k} (1-\lambda_p)^k$ & ${\tilde{L}_{sym}}$ & $K$\\
        \midrule
        \multirow{8}{*}{\rotatebox[origin=c]{90}{Infinite}} &\texttt{Scale-1}   & low-pass          & $\frac{1}{I-\beta_0 \hat{A}}$ & $\beta_0^n$  & $\tilde{A}_{sym}$ & $+\infty$\\
        &\texttt{Scale-2}    & low-pass          & $\frac{1}{(I-\beta_0 \hat{A})^2}$ & $(n+1) \beta_0^n$ & $\tilde{A}_{sym}$ & $+\infty$\\
        &\texttt{Scale-2*}     & low-pass          & $\frac{\beta_0 \hat{A}}{(I-\beta_0 \hat{A})^2}$ & $n \beta_0^n$ & ${\tilde{L}_{sym}}/{2}$ & $+\infty$\\
        &\texttt{Scale-3}      & low-pass          & $\frac{2}{(I-\beta_0 \hat{A})^3}$ & $(n+2) (n+1) \beta_0^n$ & $\tilde{A}_{sym}$ & $+\infty$\\
        &\texttt{Scale-$\alpha$}         & band-pass & $(I+\beta_0 \hat{A})^\alpha$ & $\frac{\alpha(\alpha - 1) \cdots (\alpha - n + 1)}{n!} \beta_0^n$ & $\tilde{A}_{sym}$ & $+\infty$\\
        &\texttt{Arctangent}        & high-pass          & $\text{arctan}(\beta_0 \hat{A})$ & $\frac{(-1)^{n-1}}{2n-1} \beta_0^n$ & $\tilde{L}_{sym}$ & $+\infty$\\
        &\texttt{Logarithm} & comb-pass & $\ln\frac{1}{I-\beta_0 \hat{A}}$ & $\frac{\beta_0^n}{n}$ & ${\tilde{L}_{sym}}/{2}$ & $+\infty$\\
        &\texttt{Katz} & comb-pass & $[(I-\beta_a \hat{A})^{-1}-I]/{\beta_a}$ & ${\beta_a}^{n-1}$ & ${\tilde{A}_{sym}}$ & $+\infty$\\
		\bottomrule
	\end{tabular}}
    \caption{Summary of Infinite Power Series and Polynomial Graph Filters from spectral and spatial domains. The general term of Polynomial and Infinite types could be written as spatial expression $F_{\gamma_n}(\hat{A})=\sum_{n=0}^{\text{Receptive Field}} \gamma_n \hat{A}^n$. $K$ is the maximum receptive field for Polynomial Filters. Moreover, for the Scale-$\alpha$ filter, $\alpha\in \mathbb{R}$.}
	\label{tab:filters}
\end{table}
\subsection{Graph Filter Effectiveness Analysis}
\label{Effectiveness Analysis}
As stated in Section \ref{Generalized Filter}, filter $F_\gamma(\hat{A})$ can be applied element-wise on each eigenvalue $\lambda_k$, denoted as $f_\gamma(\lambda_k)$. However, it is noticed that an efficient polynomial filter should satisfy that $f_\gamma(\lambda_k)\geq 0$~\cite{BernNET} to ensure a non-negative frequency response, which is fundamental for stable message passing and optimization within GNNs. Otherwise, the alternation of positive and negative values in the frequency response function can significantly disrupt the learning performance of GNNs by introducing instability in the message-passing process. 
Next, we choose three typical graph filters: \texttt{Scale-1}, \texttt{Logarithm}, and \texttt{Arctangent} to prove \M~ effectiveness due to space limitation:

\paragraph{{Scale-1 Graph Filter}}
For Scale-1 graph filter, aggregation matrix with $\hat{A} =\tilde{A}_{sym}$ where eigenvalues satisfy $1 = \lambda_1 \geq \lambda_2 \ldots \geq \lambda_N > -1$, it is obvious that when $0 < \beta_0 < 1/\lambda_{max}$, we have $1 \geq 1-\beta_0 \lambda_k > 0$. Therefore, $f_{\text{Scale-1}}(\lambda_k) = \frac{1}{1-\beta_0 \lambda_k} \geq 0$.


\paragraph{{Logarithm Graph Filter}}
For Logarithm graph filter where $\hat{A} = \frac{\tilde{L}_{sym}}{2}$, $1 = \lambda_1 \geq \lambda_2 \ldots \geq \lambda_N > 0$, when $0 < \beta_0 < 1/\lambda_{max}$ we have $f_{\text{Logarithm}}(\lambda_k) = \ln\frac{1}{1-\beta_0 \lambda_k} \geq 0$.

\paragraph{{Arctangent Graph Filter}}
For \texttt{Arctangent} graph filter where $\hat{A} = \tilde{L}_{sym}$, $2 = \lambda_1 \geq \lambda_2 \ldots \geq \lambda_N > 0$, let $1>f_{\text{Arctangent}}(\lambda_k) \geq 0$, we also have $0 < \beta_0 < 1/\lambda_{max}$.

\paragraph{} For other graph filters \texttt{Scale-2}, \texttt{Scale-2*}, \texttt{Scale-3}, \texttt{Scale}-$\alpha$ and \texttt{Katz}~\cite{katz}, we have also conducted corresponding analyses and all satisfy the requirements as stated in Table \ref{tab:filters} before.

\subsection{Discussion of Graph Filter Type}
\label{Discussion of Filter Type}
In this part, we discuss the graph filter type (mainly low-pass and high-pass) of \M. In spectral graph theory, the low-frequency components (smaller $\lambda$) of the eigenvalues of the graph are usually associated with the structural features of the graph; while high-frequency components (larger $\lambda$) are related to local or noise in the graph~\cite{chatterjee2024fluctuation,guo2024rethinking}. Low-pass filtering in graphs allows the low-frequency components to pass while suppressing high-frequency components, thereby highlighting the global structural features of the graph and filtering local noise. High-pass filtering emphasizes high-frequency components and suppresses low-frequency components, helping to reveal anomalies and noise in the graph~\cite{NIC_2018}.

As a consequence, we analyze the high~/~low-pass graph filter with the selected 3 kinds. Indeed, we compare our filter function with the maximum eigenvalue $\lambda_{max}$. If the graph filter's response to higher eigenvalues (i.e., high-frequency components) is relatively weaker compared to the response at the maximum eigenvalue, it is a low-pass graph filter; conversely, it is a high-pass graph filter. Specifically, we refer to the work of~\cite{GPRGNN,jin2022feature} and use division for comparison:

\paragraph{{Scale-1 Graph Filter}}
For \texttt{Scale-1} graph filter, $1 = \lambda_1 \geq \lambda_2 \ldots \geq \lambda_N > -1$, $\beta_0 \in (0, 1/{\lambda_{max}})$, the comparison with the frequency response function of the maximum eigenvalue is as follows:
\begin{equation}
\footnotesize
\begin{split}
        &  \left|{\frac{f_{\text{Scale-1}}(\lambda_k)}{f_{\text{Scale-1}}(\lambda_1)}}\right |  
        =\left| \frac{1-\beta_0}{1-\beta_0 \lambda_k} \right| \leq 1,
\end{split}\label{eq:high pass_1}
\end{equation}
so \texttt{Scale-1}  graph filter is a \textbf{low-pass graph filter}.

\paragraph{{Logarithm Graph Filter}}
For \texttt{Logarithm} graph filter $1 = \lambda_1 \geq \lambda_2 \ldots \geq \lambda_N > 0$, $\beta_0 \in (0, 1/{\lambda_{max}})$. For ease of computation, we use $\exp(\cdot)$ to rewrite the comparison as:
\begin{equation}
\footnotesize
\begin{split}
        &  \left|{\frac{\exp(f_{\text{Logarithm}}(\lambda_k))}{\exp(f_{\text{Logarithm}}(\lambda_1))}}\right |  
        / \left|\frac{\exp(\lambda_k)}{\exp(\lambda_1)} \right |
        = \frac{1-\beta_0}{1-\beta_0 \lambda_k} \frac{1}{e^{\lambda_k-1}}.
\end{split}\label{eq:high pass_2}
\end{equation}
We set $\Gamma(\lambda_k) = (1-\beta_0 \lambda_k) e^{\lambda_k-1}-(1-\beta_0)$, $\Gamma^{'}(\lambda_k) = (1-\beta_0 \lambda_k-\beta_0) e^{\lambda_k-1}$. It is noticed that $\Gamma(1)=0$, $\Gamma^{'}(1)=1-2 \beta_0$, $\Gamma^{'}(0)>0$. After analyzing the monotonicity of $\Gamma(\lambda_k)$, when ${1}/{2} \leq \beta_0 < {1}/{\lambda_{max}}$, $\forall \lambda_k \in (0,1],\Gamma(\lambda_k) \geq 0$, \texttt{Logarithm} graph filter becomes a \textbf{low-pass graph filter}. Besides, when $0 < \beta_0 < {1}/{2}$, $\forall \lambda_k \in (0,1],\Gamma(\lambda_k) < 0$ and \texttt{Logarithm} graph filter becomes a \textbf{high-pass graph filter}.

\paragraph{{Arctangent Graph Filter}}
For Arctangent graph filter, $1 = \lambda_1 \geq \lambda_2 \ldots \geq \lambda_N > 0$, $\beta_0 \in (0, 1/{\lambda_{max}})$, the comparison is as follows:
\begin{equation}
\footnotesize
\begin{split}
        &  \left|{\frac{f_{\text{Arctangent}}(\lambda_k)}{f_{\text{Arctangent}}(\lambda_1)}}\right |  
        / \left|\frac{\lambda_k}{\lambda_1} \right |
        = \frac{\text{arctan}(\beta_0 \lambda_k)}{\lambda_k \cdot \text{arctan}(\beta_0)} .
\end{split}\label{eq:high pass_3}
\end{equation}
Here we set $\Gamma(\lambda_k) = \text{arctan}(\beta_0 \lambda_k) - \lambda_k \text{arctan}(\beta_0)$, $\Gamma^{'}(\lambda_k) = \frac{\beta_0}{1+\beta^2_0 \lambda^2_k}-\text{arctan}(\beta_0)$ for simplicity. We notice that $\Gamma(0)=0$, $\Gamma(1)=0$ and $\Gamma^{'}(\lambda_k)$ is monotonically decreasing. Thus when $\forall \lambda_k \in (0,1],\Gamma(\lambda_k) \geq 0$, \texttt{Arctangent} graph filter becomes a \textbf{high-pass graph filter}.

\subsection{Relations Between Power Series Filters and Previous Work}
\label{sec: Relations Between Power Series Filter and Previous Work} 
In this section, we compare \M~ ($F_\gamma(\hat{A}) = \sum_{n=0}^{+\infty} \gamma_n \hat{A}^n$) with other graph filters. As shown in Table \ref{tab:filters}, we set $\gamma_n=\beta_a^n$ to realize Katz filter~\cite{katz} and $\gamma_n=\alpha (1-\alpha)^n$ to realize APPNP~\cite{APPNP}. Other polynomial graph filters can also be extended by \M. Taking Res-GCN~\cite{ResGCN} for example, we consider the residual connection and ignore the learnable weights and activation function of GNN layers from $1$ to $K-1$ following SGC~\cite{SGC}. Then, the node representation matrix from Eq.(\ref{eq:MPNN-aggretation}) could be rewritten as: $
        H^{(l+1)} = \sigma \bigl(\hat{A} (H^{(0)}+\beta_a H^{(1)} + \cdots +\beta_a^K H^{(K)}) W^{(l)} \bigl),  \noindent
$
where the shrink coefficient $\beta_a$ is used by concatenating or adding each layer. Thus, we get a more general expression where the coefficients follow the binomial theorem $
        H^{(l+1)} = \beta_a^K (\sum_{k=0}^{K} \binom{K}{k}\hat{A}^k) X = (\beta_a \hat{A} + \beta_a I)^K X.
$
Thus, Res-GCN also belongs to \M~ with $\gamma_n = \binom{K}{n}$. Besides, GPR-GNN~\cite{GPRGNN} proposes a general formulation of the polynomial graph filter which can be regarded as a limited form of GPFN by constraining the infinite polynomial to a certain range. Note that these polynomial graph filters have an upper bound of aggregation horizon as the GNN layer is fixed, restricting their abilities to capture long-range dependency.

\section{Experiments}
\label{experiments}
In this section, we conduct a series of experiments on three datasets to answer the following research questions:
\begin{itemize}[leftmargin=*]
    \item \textbf{RQ1}: Does \M~outperform the state-of-the-art baselines under the highly sparse graph scenario? 
    \item \textbf{RQ2}: Is our \M~ a flexible graph filter framework? And what are the effects of different power-series graph filters?
    \item \textbf{RQ3}: How sensitive is our GPFN to hyper-parameters $\beta_0$?
    \item \textbf{RQ4}: Can our infinite graph filters learn long-range information at the shallow layer and alleviate over-smoothness?
    \item \textbf{RQ5}: How can our \M~ provide interpretability on the nature graph or other fields?
\end{itemize}
\begin{table}[t]
  \centering
  \tiny
   \setlength\tabcolsep{2pt}
    \begin{threeparttable}
\begin{tabular}{c|c|cccccc}
    \hline \textbf{Dataset}& \textbf{Type} & \textbf{Nodes} &  \textbf{Edges} & \textbf{$L_{sym}$ Eigenvalues} & \textbf{Classes} &
     \textbf{Sparsity}&
\textbf{Train/Valid/Test} \\ \hline
     Cora & Binary & 2,708 & 5,429 & [0,1.999] & 7 &  99.93\% & 140/500/1,000 \\
     Citeseer & Binary & 3,327 & 4,732 & [0,1.502] & 6 & 99.96\% & 120/500/1,000 \\
     AmaComp & Binary & 13,752 & 245,861 & [0,1.596] & 10  & 99.87\% & 400/500/12,852 \\ 
     \hline
\end{tabular}
    \end{threeparttable}
  \caption{The statistics of datasets.}
  \label{tb:1}%
  \vspace{-10pt}
\end{table}
\subsection{Experimental Setup}
\subsubsection{Datasets}
In this paper, we conduct experiments on three widely used node classification datasets to assure a diverse validation, and the statistics of these datasets are summarized in Table \ref{tb:1}. 
\begin{itemize}[leftmargin=*]
\item \textbf{Cora\footnote{\label{citedata}https://github.com/kimiyoung/planetoid}}:
It is a node classification dataset that contains citation graphs, where nodes, edges, and labels in these graphs are papers, citations, and the topic of papers.
\item \textbf{Citeseer\footref{citedata}}: Similar to the Cora dataset, Citeseer is a citation graph-based node classification dataset.
\item \textbf{AmaComp\footnote{https://github.com/shchur/gnn-benchmark}}:
It is a node classification dataset that contains product co-purchase graphs, where nodes, edges, and labels in these graphs are Amazon products, co-purchase relations, and the category of products. Compared to Cora and Citeseer, AmaComp is denser and larger.

\end{itemize}
Moreover, to verify the capability of our GPFN in extracting information on the sparse graph, we test our methods and baselines on sparse datasets, \emph{i.e.}, we randomly remove the masking ratio (MR) percentage of edges from original datasets before training.


\subsubsection{Baselines}

We compare our \M~ with $18$ baselines, from four MR categories for comprehensive experiments: i) Non-graph filter-based methods: \textbf{MLP}~\cite{mlp} and \textbf{LP}~\cite{zhuѓ2002learning}. ii) Monomial graph filter-based methods: \textbf{GCN}~\cite{GCN}, \textbf{GAT}~\cite{GAT}, \textbf{GIN}~\cite{gin}, \textbf{AGE}~\cite{AGE}, \textbf{GCN-SGC} and \textbf{GAT-SGC}~\cite{SGC}. iii) Polynomial graph filter-based methods: \textbf{ChebGCN}~\cite{chebnet}, \textbf{GPR-GNN}~\cite{GPRGNN},
\textbf{APPNP}~\cite{APPNP}, \textbf{BernNet}~\cite{BernNET}, 
\textbf{GRAND}~\cite{2022GRAND},
\textbf{GCNII}~\cite{chen2020simple},
\textbf{ADC}~\cite{NEURIPS2021_c42af2fa},
\textbf{DGC}~\cite{wang2021dissecting},
\textbf{D$^2$PT}~\cite{10.1145/3580305.3599410}, \textbf{HiGNN}~\cite{HiGCN}, \textbf{HiD-GCN}~\cite{li2023generalized} \textbf{Res-GCN} and \textbf{Res-GAT}~\cite{ResGCN}. Detailed description of baselines can be referred to in Appendix C.
\begin{table*}[htbp]
  \centering
  \setlength\tabcolsep{3.5pt}
  
  \begin{threeparttable}
    \resizebox{1.0\linewidth}{!}{\begin{tabular}{c|l|cccc|cccc|cccc}
    \hline
    \hline
   \multicolumn{2}{c|}{Datasets} & \multicolumn{4}{c|}{\textbf{Cora}} & \multicolumn{4}{c|}{\textbf{Citeseer}} & \multicolumn{4}{c}{\textbf{AmaComp}} \\
    \cline{1-14}          
    \multicolumn{2}{c|}{\textbf{MR}} & 0\%   & 30\%   & 60\%   & 90\%   & 0\% & 30\% & 60\% & 90\% & 0\% & 30\% & 60\% & 90\%\\
    \hline
    \multirow{18}{*}{\rotatebox[origin=c]{90}{Baselines}} & MLP & 49.97$\pm$2.30 & 50.40$\pm$2.23 & 51.49$\pm$1.47 & 47.78$\pm$1.91 & 52.18$\pm$2.15 & 48.46$\pm$2.19 & 49.64$\pm$3.79 & 52.90$\pm$2.21 & 67.87$\pm$1.22 & 66.23$\pm$0.80 & 67.73$\pm$1.62 & 68.13$\pm$1.88\\
    & LP & 71.80$\pm$1.02 & 58.29$\pm$1.45 & 53.41$\pm$2.27 & 50.27$\pm$3.46 & 51.30$\pm$1.43 & 50.08$\pm$1.62 & 48.32$\pm$2.04 & 46.42$\pm$2.48 & 74.84$\pm$1.52 & 71.29$\pm$1.65 & 70.38$\pm$1.88 & 69.10$\pm$2.01\\
    \cline{2-14}
    & GCN & 75.73$\pm$1.86 & 67.88$\pm$1.91 & 62.67$\pm$2.77 & 54.39$\pm$2.72 & 66.37$\pm$1.28 & 61.82$\pm$1.54 & 62.03$\pm$1.08 & 56.60$\pm$1.86 & 80.77$\pm$0.44 & 79.44$\pm$1.06 & 78.73$\pm$1.02 & 73.62$\pm$1.27\\
    & GAT & 76.86$\pm$1.41 & 73.34$\pm$2.38 & 65.07$\pm$1.52 & 53.91$\pm$3.17 & 66.30$\pm$0.51 & 63.73$\pm$2.26 & 60.25$\pm$0.96 & 54.33$\pm$1.48 & 74.61$\pm$3.31 & 73.51$\pm$3.54 & 73.48$\pm$2.06 & 74.15$\pm$0.72\\
    & GIN & 74.16$\pm$2.76 & 65.95$\pm$4.99 & 58.69$\pm$4.65 & 50.71$\pm$5.34 & 65.87$\pm$2.26 & 60.80$\pm$2.34 & 59.25$\pm$2.93 & 54.60$\pm$2.66 & 74.35$\pm$2.25 & 72.59$\pm$4.31 & 70.04$\pm$5.94 & 68.65$\pm$5.56\\
    & AGE & 67.11$\pm$1.70 & 65.79$\pm$1.99 & 59.96$\pm$1.16 & 56.31$\pm$1.97 & 67.11$\pm$1.70 & 65.79$\pm$1.99 & 59.96$\pm$1.15 & 55.31$\pm$1.97 & 76.53$\pm$1.46 & 77.70$\pm$2.23 & 76.14$\pm$0.67 & 73.84$\pm$0.68\\
    & GCN-SGC & 79.35$\pm$1.44 & 72.33$\pm$1.86 & 63.58$\pm$1.33 & 57.18$\pm$2.02 & 63.64$\pm$1.18 & 63.10$\pm$1.77 & 62.50$\pm$1.00 & 55.13$\pm$1.52 & 72.78$\pm$1.48 & 70.74$\pm$0.50 & 73.48$\pm$0.93 & 71.77$\pm$0.28\\
    & GAT-SGC & 75.10$\pm$2.24 & 66.69$\pm$2.78 & 58.48$\pm$3.11 & 51.09$\pm$3.65 & 62.65$\pm$2.52 & 64.46$\pm$2.89 & 64.02$\pm$1.76 & 52.13$\pm$3.01 & 72.74$\pm$5.11 & 79.34$\pm$1.61 & 70.56$\pm$1.43 & 66.16$\pm$3.63\\
    \cline{2-14}
    & ChebGCN & 76.97$\pm$1.74 & 73.40$\pm$2.16 & 63.02$\pm$2.03 & 50.61$\pm$2.80 & 65.13$\pm$1.88 & 62.92$\pm$1.72 & 59.00$\pm$2.38 & 49.95$\pm$1.94 & 81.09$\pm$0.43 & 80.44$\pm$0.64 & 77.22$\pm$0.91 & 73.62$\pm$1.13\\
    & GPRGNN & 79.54$\pm$1.37 & 76.05$\pm$1.48 & 65.59$\pm$2.98 & 54.30$\pm$3.41 & 68.55$\pm$0.94 & 64.89$\pm$2.78 & 61.92$\pm$1.32 & 52.85$\pm$1.25 & 82.42$\pm$1.28 & 81.43$\pm$0.73 & 80.43$\pm$0.73 & 77.11$\pm$1.51\\
    & APPNP & 76.90$\pm$1.42 & 74.01$\pm$2.57 & 63.81$\pm$2.27 & 51.84$\pm$3.96 & 68.69$\pm$1.29 & 64.03$\pm$1.49 & 61.93$\pm$1.15 & 53.75$\pm$1.99 & 79.60$\pm$0.63 & 79.67$\pm$1.12 & 78.56$\pm$1.54 & 76.39$\pm$1.54\\
    & RES-GCN & 76.93$\pm$1.38 & 76.42$\pm$1.55 & \textcolor{cyan}{69.34$\pm$1.93} & 49.42$\pm$2.30 & 67.28$\pm$1.10 & 64.53$\pm$1.25 & 62.44$\pm$1.56 & 51.80$\pm$1.98 & 77.53$\pm$1.53 & 75.31$\pm$1.12 & 74.63$\pm$1.23 & 73.83$\pm$1.19\\
    & RES-GAT & 74.06$\pm$0.94 & 71.12$\pm$1.35 & 65.06$\pm$1.70 & 53.95$\pm$2.02 & 67.51$\pm$1.64 & 63.88$\pm$1.85 & 62.63$\pm$2.09 & 49.11$\pm$2.57 & 72.03$\pm$1.27 & 70.15$\pm$1.50 & 72.03$\pm$1.45 & 68.22$\pm$2.73\\
    & BernNet & 79.97$\pm$2.48 & 72.56$\pm$1.79 & 66.48$\pm$1.80 & 48.00$\pm$3.09 & 74.35$\pm$0.53 & 68.62$\pm$0.74 & 61.04$\pm$0.93 & 47.71$\pm$1.60 & 82.03$\pm$1.17 & 81.34$\pm$1.35 & 75.69$\pm$1.55 & 70.78$\pm$2.06\\
    & GCNII & 73.53$\pm$2.34 & 68.18$\pm$2.78 & 61.80$\pm$4.33 & 51.26$\pm$1.04 & 63.86$\pm$1.57 & 62.56$\pm$0.78 & 57.03$\pm$1.34 & 53.68$\pm$0.52 & 70.39$\pm$3.02 & 70.20$\pm$2.57 & 69.80$\pm$1.78 & 65.76$\pm$2.33\\
    & ADC & 78.16$\pm$0.84 & 73.76$\pm$1.18 & 63.72$\pm$1.27 & 51.28$\pm$1.67 & 72.18$\pm$1.43 & 68.79$\pm$0.62 & 62.72$\pm$1.29 & 52.61$\pm$2.95 & 79.55$\pm$1.34 & 79.35$\pm$1.88 & 78.49$\pm$1.34 & 72.29$\pm$2.34\\
    & DGC & 79.85$\pm$1.14 & 75.78$\pm$2.93 & 62.75$\pm$1.36 & 50.45$\pm$1.54 & 73.45$\pm$0.91 & 69.32$\pm$1.62 & 62.07$\pm$2.54 & 55.60$\pm$1.80 & 81.43$\pm$0.71 & 80.98$\pm$1.49 & 80.61$\pm$3.01 & 74.88$\pm$2.34\\
    & GRAND & 79.44$\pm$1.89 & 74.23$\pm$2.01 & 64.33$\pm$2.45 & 52.09$\pm$2.70 & 74.36$\pm$1.03 & 69.98$\pm$1.69 & 63.20$\pm$1.54 & 55.41$\pm$2.48 & \textcolor{cyan}{83.57$\pm$2.44} & 81.42$\pm$2.60 & 80.11$\pm$2.57 & 74.39$\pm$2.88\\
    & D$^2$PT & 79.31$\pm$1.22 & 74.47$\pm$1.78 & 64.87$\pm$2.38 & 51.48$\pm$3.44 & 75.28$\pm$1.94 & 69.32$\pm$1.97 & \textcolor{cyan}{64.77$\pm$2.01} & 56.82$\pm$2.73 & 82.80$\pm$1.88 & 80.92$\pm$1.92 & 79.20$\pm$2.03 & 77.48$\pm$2.46\\
    & HiGNN & 80.03$\pm$1.48 & 76.14$\pm$1.73 & 64.38$\pm$1.93 & 50.26$\pm$2.08 & 74.88$\pm$1.10 & 69.52$\pm$1.31 & 63.92$\pm$2.29 & 53.44$\pm$1.94 & 81.88$\pm$1.57 & 81.34$\pm$1.43 & 79.99$\pm$1.60 & 75.04$\pm$2.39\\
    & HiD-GCN & 78.42$\pm$1.57 & 76.20$\pm$1.61 & 64.62$\pm$1.80 & 52.39$\pm$2.74 & 74.65$\pm$1.03 & 68.77$\pm$1.41 & 64.03$\pm$1.72 & 55.49$\pm$2.39 & 80.06$\pm$1.91 & 79.94$\pm$2.94 & 75.32$\pm$2.30 & 73.28$\pm$2.47\\
    \hline
    \hline
    \multirow{10}{*}{\rotatebox[origin=c]{90}{GPFN}} & GCN-\texttt{S1} & 80.15$\pm$1.32 & \redfont{\textbf{76.53$\pm$1.23}} & 68.01$\pm$1.85 & 59.33$\pm$1.76 & \textcolor{cyan}{76.85$\pm$1.32} & \textcolor{cyan}{72.52$\pm$1.53} & 64.76$\pm$1.75 & 59.33$\pm$1.71 & \textcolor{red}{\textbf{83.90$\pm$1.37}} & \redfont{\textbf{83.61$\pm$1.80}} & \redfont{\textbf{83.23$\pm$2.24}} & \textcolor{cyan}{78.98$\pm$0.65}\\
    & GCN-\texttt{S2} & \textcolor{cyan}{80.33$\pm$1.88} & 76.42$\pm$1.15 & 68.09$\pm$2.16 & 54.74$\pm$1.74 & \redfont{\textbf{78.33$\pm$1.73}}& \redfont{\textbf{74.42$\pm$1.26}} & \redfont{\textbf{66.09$\pm$1.08}} & 56.74$\pm$1.72 & 81.66$\pm$1.75 & 80.87$\pm$0.31 & \textcolor{cyan}{81.14$\pm$0.34} & \redfont{\textbf{79.79$\pm$2.30}}\\
    & GCN-\texttt{S3} & 79.46$\pm$1.02 & \textcolor{cyan}{76.45$\pm$1.17} & 69.07$\pm$1.84 & 55.56$\pm$1.97 & 71.69$\pm$1.28 & 69.47$\pm$1.33 & 63.44$\pm$1.40 & \redfont{\textbf{60.03$\pm$2.05}} & 76.25$\pm$1.28 & 77.30$\pm$1.53 & 74.10$\pm$1.74 & 73.29$\pm$1.77\\
    & GCN-\texttt{Log} & 79.61$\pm$1.38 & 74.06$\pm$1.76 & 67.65$\pm$2.22 & 58.88$\pm$1.98 & 69.59$\pm$1.47 & 67.53$\pm$1.36 & 60.55$\pm$2.46 & 59.09$\pm$1.63 & 81.37$\pm$1.27 & \textcolor{cyan}{82.46$\pm$2.34} & 78.01$\pm$2.40 & 77.23$\pm$1.90\\
    & GCN-\texttt{Katz} & \redfont{\textbf{80.77$\pm$1.25}} & 75.76$\pm$2.67 & \redfont{\textbf{69.51$\pm$1.70}} & \redfont{\textbf{64.25$\pm$1.39}} & 69.40$\pm$1.39 & 66.51$\pm$1.70 & 64.71$\pm$1.32 & 57.35$\pm$1.39 & 71.07$\pm$2.18 & 76.76$\pm$2.67 & 74.72$\pm$1.54 & 75.79$\pm$0.37\\
    \cline{2-14}
    & GAT-\texttt{S1} & 75.95$\pm$1.10 & 72.86$\pm$1.46 & 66.76$\pm$1.58 & \textcolor{cyan}{62.35$\pm$1.84} & 65.10$\pm$0.78 & 63.18$\pm$1.40 & 63.00$\pm$1.88 & 58.92$\pm$1.79 & 78.64$\pm$1.92 & 75.11$\pm$1.53 & 74.72$\pm$1.36 & 64.49$\pm$1.96\\
    
    & GAT-\texttt{S2} & 79.12$\pm$0.86 & 72.88$\pm$1.22 & 66.02$\pm$1.49 & 58.82$\pm$1.44 &76.46$\pm$0.86 & 70.57$\pm$1.38 & 61.21$\pm$1.82 & 55.01$\pm$1.94 & 73.69$\pm$0.76 & 75.95$\pm$1.21 & 69.04$\pm$1.40 & 62.06$\pm$1.68\\
    
    & GAT-\texttt{S3} & 75.70$\pm$0.97 & 72.65$\pm$1.01 & 65.29$\pm$1.44 & 54.86$\pm$2.30 & 70.98$\pm$1.42 & 69.16$\pm$1.87 & 60.53$\pm$1.99 & \textcolor{cyan}{59.59$\pm$1.91} & 74.88$\pm$1.06 & 75.51$\pm$1.34 & 73.22$\pm$1.37 & 68.75$\pm$1.88\\
    & GAT-\texttt{Log} & 75.95$\pm$1.22 & 72.86$\pm$1.40 & 66.76$\pm$1.68 & 62.35$\pm$1.92 & 65.10$\pm$0.94 & 63.18$\pm$1.27 & 63.00$\pm$1.45 & 58.92$\pm$1.81 & 78.64$\pm$1.22 & 75.11$\pm$1.87 & 74.72$\pm$1.94 & 64.49$\pm$2.03\\
    & GAT-\texttt{Katz} & 79.36$\pm$1.80 & 75.05$\pm$1.82 & 68.21$\pm$1.13 & 60.39$\pm$1.70 & 71.89$\pm$1.51 & 68.21$\pm$1.13 & 63.10$\pm$1.12 & 59.39$\pm$1.70 & 77.36$\pm$1.80 & 73.05$\pm$1.82 & 75.93$\pm$2.90 & 71.16$\pm$1.79\\
    \hline
    \hline
    \end{tabular}}%
    \end{threeparttable}
    \caption{Node classification accuracy (in percent ± standard deviation) comparison on all datasets under different edge masking ratios (MR). \redfont{\textbf{Bold in redfont}}: best performance, \textcolor{cyan}{Bluefont}: second best performance.} 
    \label{tab:cor}
  \vspace{-10pt}
\end{table*}%

\subsubsection{Hyper-parameter Settings}


The learnable parameters of our model are optimized for $200$ epochs by the Adam optimizer~\cite{Adam} with a learning rate of $0.002$ and a weight decay of $0.005$. Besides, we employ the early-stopping strategy with patience equal to $20$ to avoid over-fitting. To show the flexible design of our GPFN, we incorporate the \texttt{Scale-1}, \texttt{Scale-2}, \texttt{Scale-3}, \texttt{Logarithm}, and \texttt{Katz} filters with the GCN and GAT framework under the blend factor $\beta_0=0.8$, namely GCN- or GAT-\texttt{S1}, -\texttt{S2}, -\texttt{S3}, -\texttt{Log}, and -\texttt{Katz} for short. The GNNs in \M~ for all variants are two-layered with the hidden units to 16. Moreover, hyper-parameter settings of baselines can be referred to in Appendix D. Finally, for a fair comparison, we repeat all experiments $10$ times with randomly initialed parameters and show the average value with the standard deviation in our paper and the statistically significant results are $p < 0.05$.



\subsubsection{Experimental Settings}
Our methods and baselines are implemented by the deep learning framework PyTorch 1.9.0~\cite{paszke2019pytorch} with the programming language Python 3.8. All of the experiments are conducted on a Ubuntu server with the NVIDIA Tesla V100 GPU and the Intel(R) Xeon(R) CPU. Baselines are all implemented using their official source codes.
\subsection{Main Results~(RQ1)}
To answer RQ1, we conduct experiments and report results of node classification accuracy on the Cora, Citeseer, and AmaComp datasets, as illustrated in Table~\ref{tab:cor}. From the reported accuracy, we can find the following observations:

\textbf{Comparison of graph filters with polynomial filters.} The results of graph filter-based methods such as GCN demonstrate that using graph filters in GNN can significantly improve the performance of node classification compared to non-graph filter-based methods such as MLP, demonstrating the importance of graph structure. In addition, while recent monomial graph filter-based works (\emph{e.g.}, AGE, GCN-SGC, GAT-SGC) attempt to design low-pass filters, improvements are gained compared with GCN and GAT because of the neglecting of high-order structure information in propagating graph signals. With the guidance of polynomial graph filters, GPR-GNN, BernNet, HiGNN, etc. surpass previous monomial graph filter-based methods by enlarging receptive field.




\textbf{Consistent Performance Superiority.} The performance test of baselines consistently shows that our proposed power series filters enhanced GCN and GAT outperform almost baselines especially in sparse graph settings (\emph{i.e.}, large edge masking ratio) gained from 0.17\%-7.07\% on Cora, 1.32\%-4.44\% on Citeseer and 0.33\%-2.8\% on Amaphoto. This highlights the effectiveness of our methods in aggregating long-range information via filtering high-frequency noise from the sparse graph. In addition, we found that GPFN was the most stable in highly sparse scenes, indicating that our model is more robust because GPFN effectively alleviates the sparse connection problem and focuses more on long-range information in which this learned global knowledge is essential for leading to improved classification accuracy. However, it is noticed that GPFN with GAT frameworks' effects are not satisfying especially when $\text{MR}$ is small, and we argue that it is because the learned attention is essentially a reweighted repair of the GPFN filter, this will interfere with the filter function $f_\gamma(\cdot)$ of GPFN from spatial view. From the perspective of the spatial domain, we argue that the learned attention will bring a lot of redundant information, and when the graph is not too sparse, it will cause the risk of overfitting. 


\subsection{Flexibility Analysis~(RQ2)}
To show the flexibility of our GPFN in incorporating different graph filters, we show the performance of GPFN variants in Table~\ref{tab:cor}. According to the results of these variants, we have the following observations. First, we can notice that our method in different filter settings can achieve better performance compared to baselines under the sparse graph, especially when $\text{MR} \geq 0.6$. This finding verifies the effectiveness of our flexible design in the sparse graph-based node classification. Moreover, for GPFN, other graph filters work better compared to the \texttt{Log}, because the \texttt{Log} passes through more high-frequency signals when $\beta_0=0.8$. However, different filters achieve different effects on different datasets, as per Wolpert’s 'No Free Lunch' theorem~\cite{lunch}. Besides, the improvements become inapparent when handling AmaComp at high MR. We conjecture that in these cases, previous GNN filter can aggregate sufficient information from message-passing framework, and messages introduced by the receptive field expansion become unnecessary.


\begin{figure}[htb]
  \centering
  \includegraphics[scale=0.25]{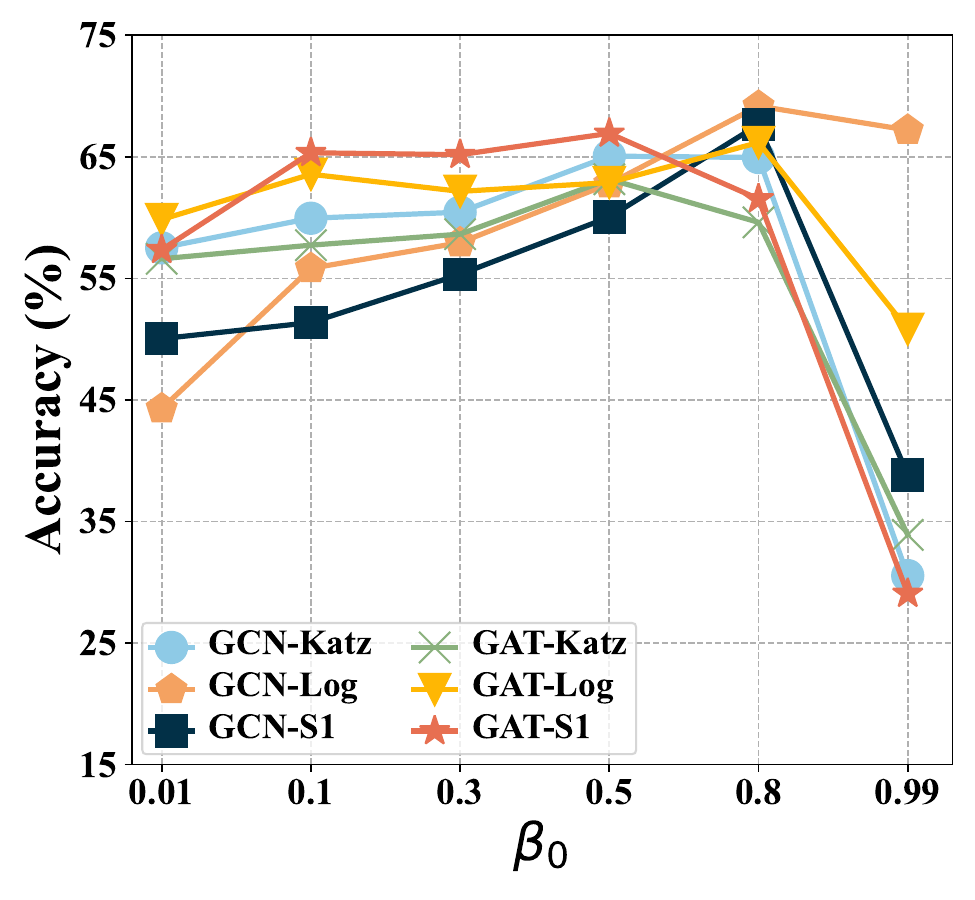}
  \includegraphics[scale=0.25]{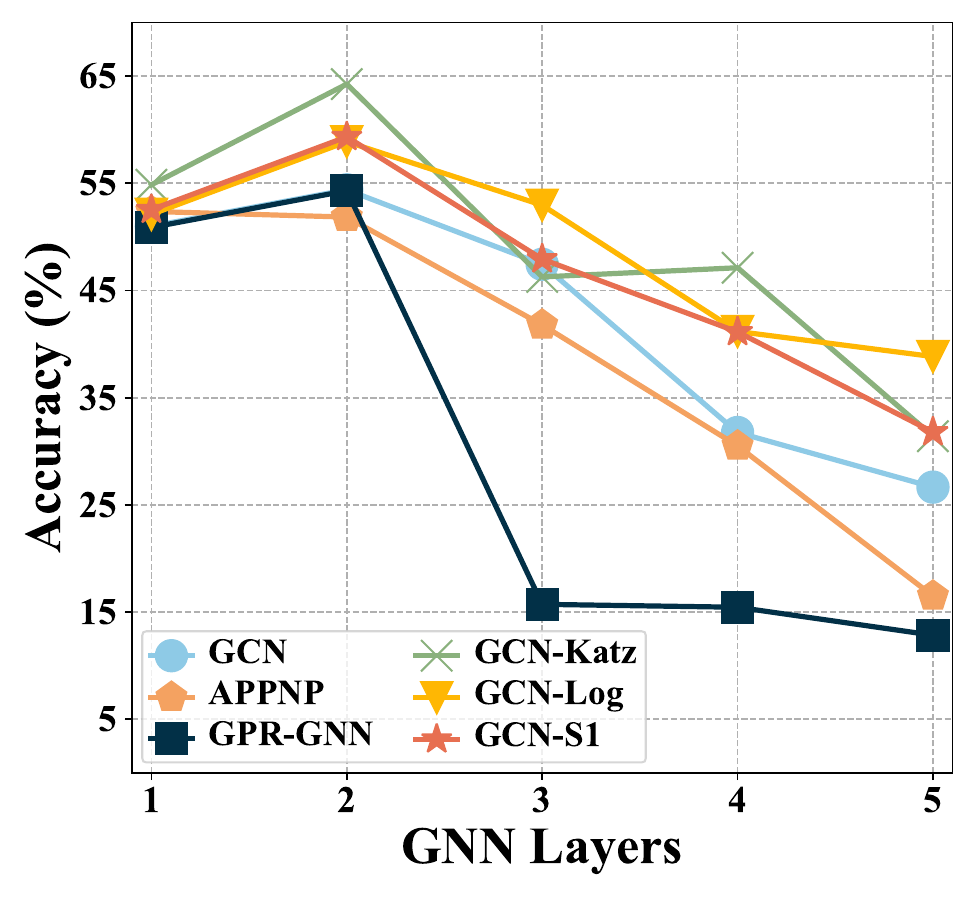}
  \vspace{-5pt}
  \caption{(\textbf{Left}.) Hyper-parameter study with the blend factor $\beta_0$ on Cora from 0.01 to 0.99 when $\text{MR}=0.60$. (\textbf{Right}.) Long-range study with the GNN layers from 1 to 5 on Cora when $\text{MR}=0.90$.
  }
   \label{fig:hyper_study}
\end{figure}

\subsection{Hyper-parameter Study~(RQ3)}

In this section, we concentrate on evaluating the influence of different hyper-parameters on GPFN for RQ3. Specifically, we perform a series analysis of blend factor $\beta_0$ from the list $[0.01,0.1,0.3,0.5,0.8,0.99]$ to design our infinite graph filter. The left part of Figure~\ref{fig:hyper_study} depicts the best performance for different filters is respectively achieved when $\beta_0=0.5$ and $\beta_0=0.8$, emphasizing the significance of scaling blend factor for learning the sparse graph. We notice performance for \texttt{Katz} and \texttt{S1} drop rapidly when $\beta_0=0.99$, while filter Log still performs well. We analyze that \texttt{Katz} and \texttt{S1} use $\beta_0^n$ as $\gamma_n$ while \texttt{Log} adapts $\gamma_n=\frac{\beta_0^n}{n}$. When $\beta_0$ is close to $1$, the weights of every hop in the aggregation matrix are nearly identical from the spatial view, and it can also be analyzed in the spectral domain because the filter function $F_{\gamma}(\cdot)$ allows more high-frequency signals. 



\subsection{Long-Range Study~(RQ4)}
In this section, to verify the effectiveness of learning long-range dependencies of GPFN, we vary the GNN layers number from 1 to 5 when $\text{MR}=0.60$ of GCN, APPNP, GPR-GNN, and GCN-\texttt{Katz} on Cora. Theoretically, the spatial receptive field of GNNs expands as layers increase, but the over-smoothing problem that arises in deeper networks makes training harder. As shown in Figure~\ref{fig:hyper_study}, 
the best performance of all methods is achieved with the small layer due to the over-smoothing. However, our filters exhibit a slower rate of decline, which effectively counteracts the over-smoothing effect than other models. Moreover, our GPFN achieves superior performance with shallow layers compared to others, corresponding to our contribution. 

\begin{figure}[htbp]
\vspace{-0.3mm}
	\centering
	\begin{minipage}{\columnwidth}
		\centering
	\end{minipage}
	
	\begin{minipage}{0.51\columnwidth}
		\centering
		\includegraphics[width=0.99\columnwidth]{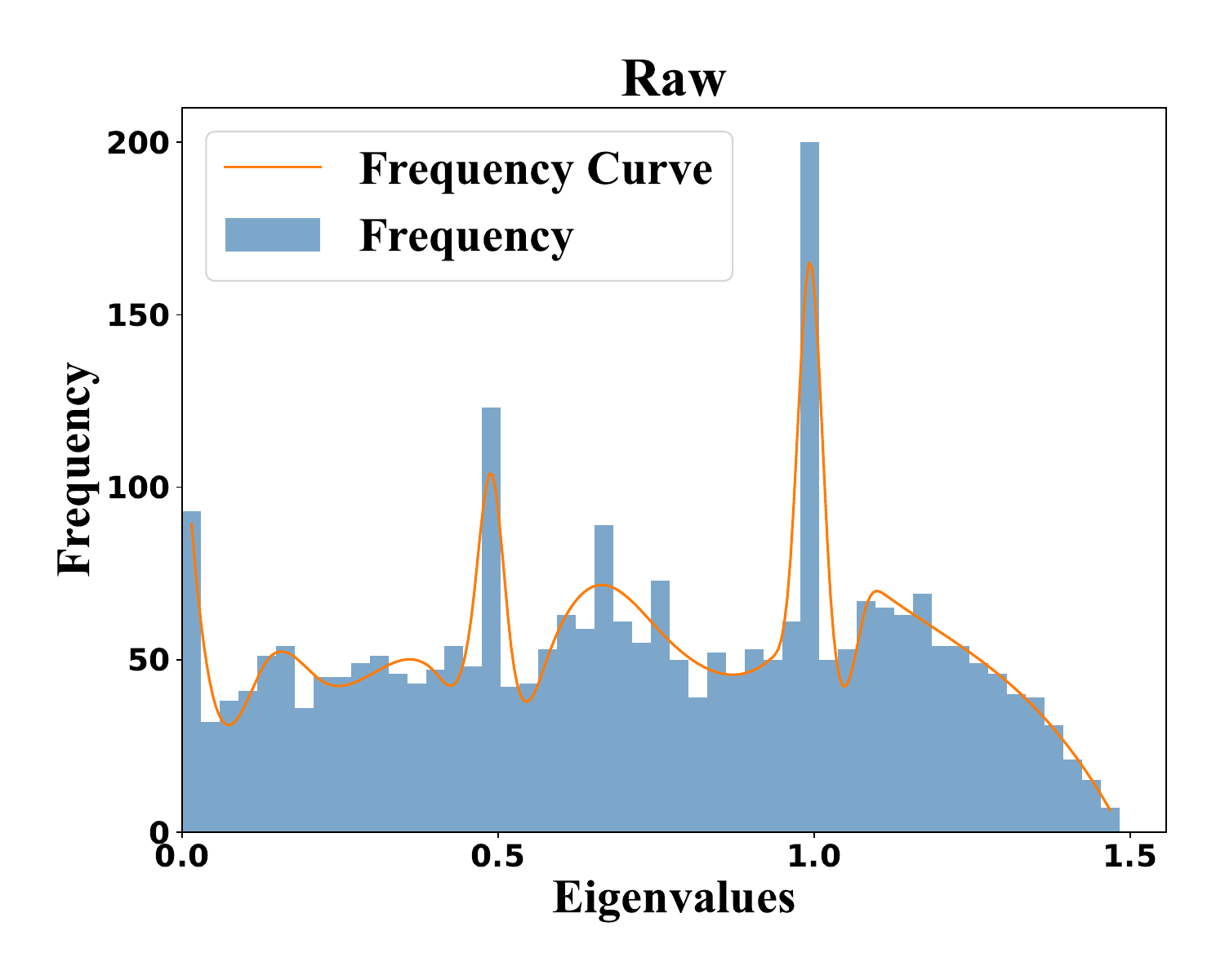}
		\vspace{-5mm}
	\end{minipage}
 \hspace{-4mm}
	\begin{minipage}{0.51\columnwidth}
		\centering
		\includegraphics[width=0.99\columnwidth]{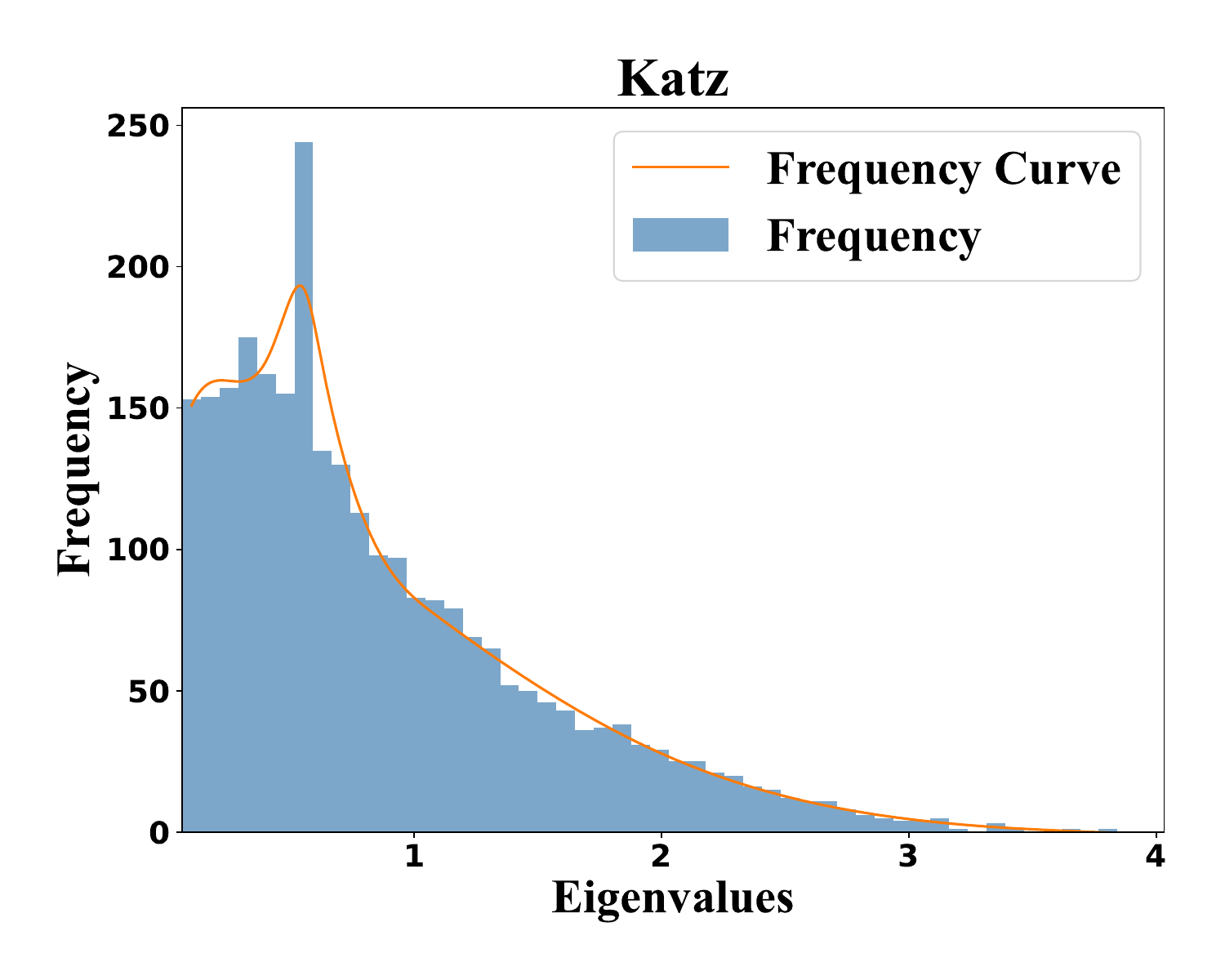}
		\vspace{-5mm}
	\end{minipage}
	\quad
	\begin{minipage}{0.51\columnwidth}
		\centering
		\includegraphics[width=0.99\columnwidth]{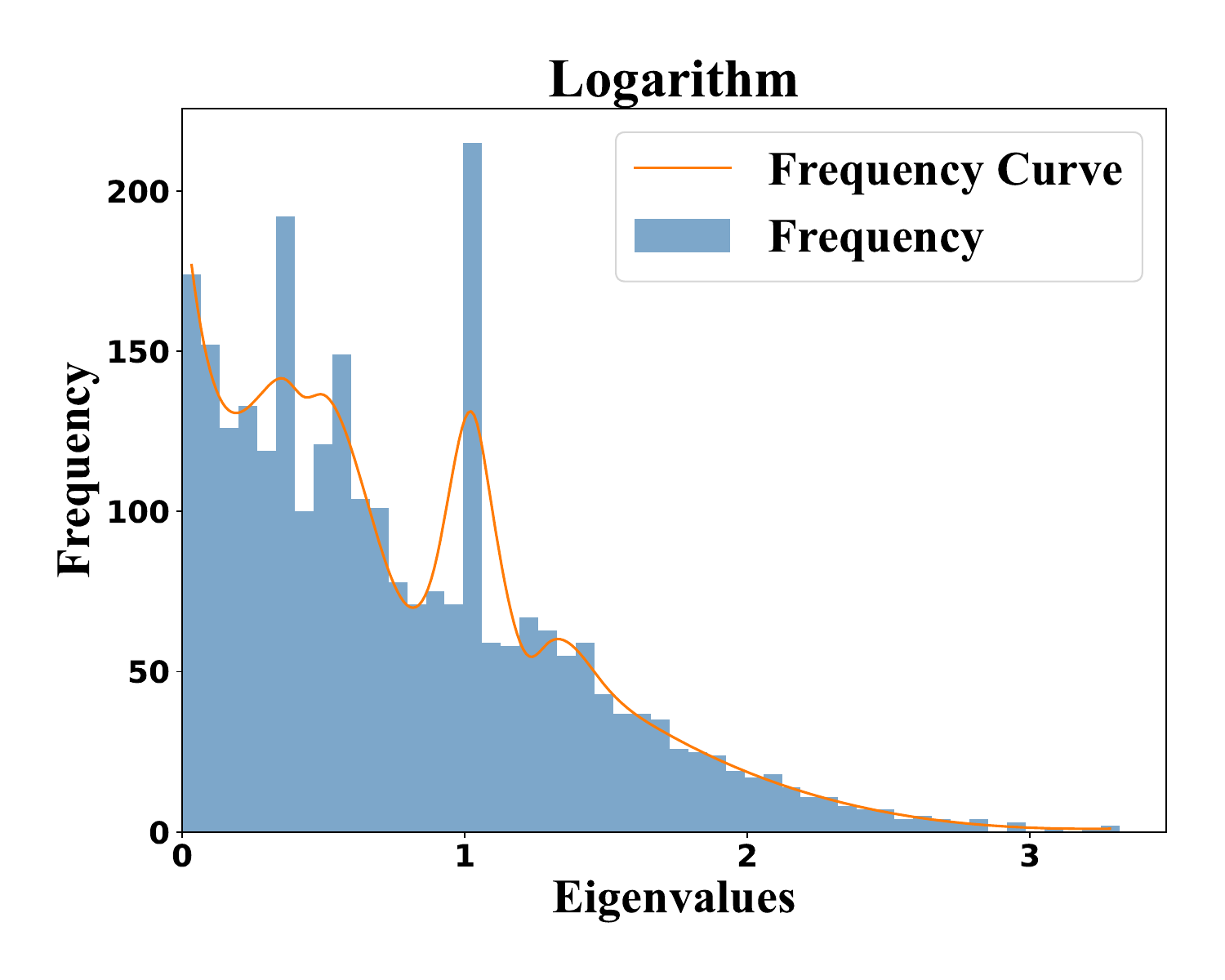}
		\vspace{-7mm}
	\end{minipage}
 \hspace{-4mm}
	\begin{minipage}{0.51\columnwidth}
		\centering
		\includegraphics[width=0.99\columnwidth]{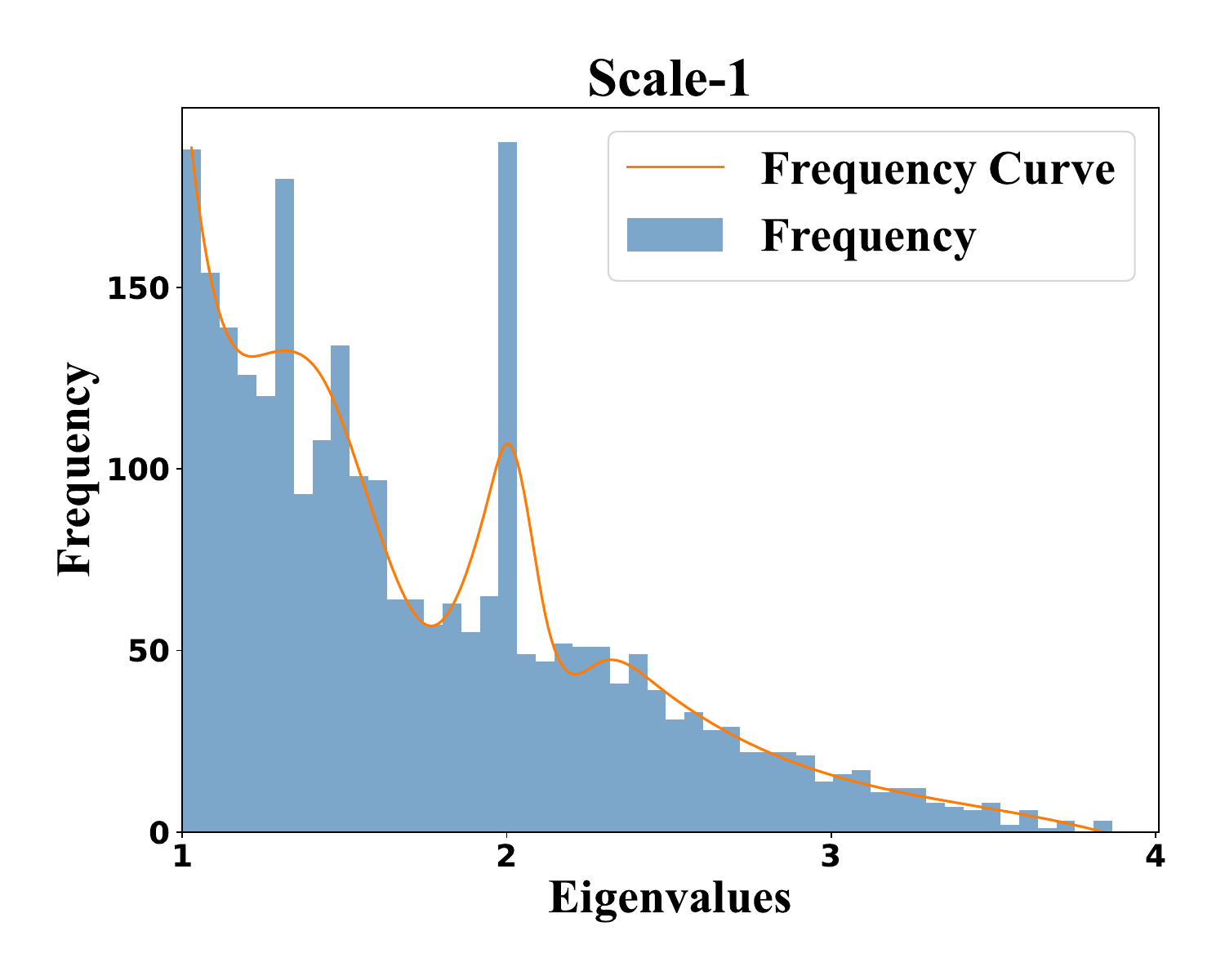}
		\vspace{-7mm}
	\end{minipage}
	  \caption{A filter study of Raw (Laplacian Aggregator), \texttt{Katz}, \texttt{Logarithm} and \texttt{Scale-1} graph filters with $\beta_0=0.8$ on Cora. Where $x$-axis is the eigenvalues and $y$-axis is the node frequency.}
  \label{fig:cora_eleg}
  \vspace{-10pt}
\end{figure}

\begin{figure}[ht]
  \centering
  \includegraphics[width=0.98\columnwidth]{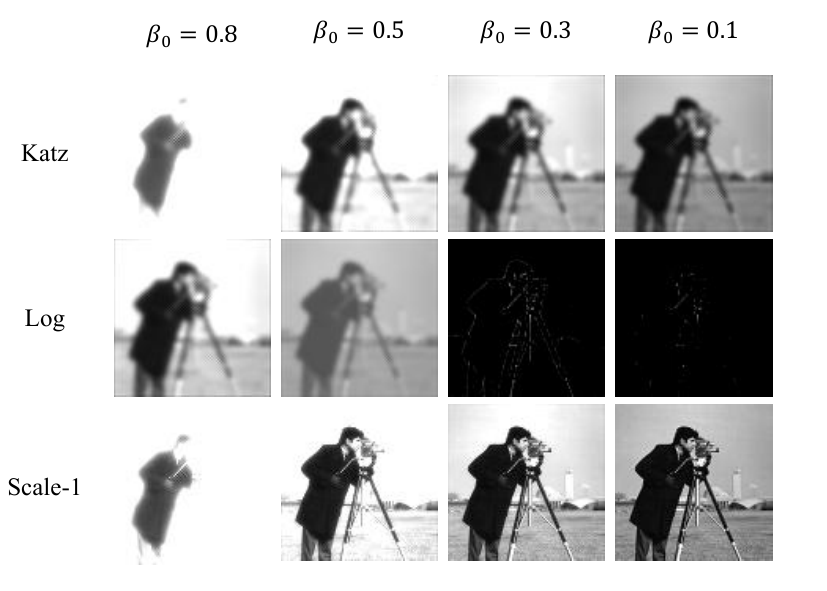}
  \caption{An input image and filtering results with \texttt{Logarithm}, \texttt{Katz}, and \texttt{Scale-1} graph filters. We also vary $\beta_0$ from 0.8 to 0.1.}
  \label{fig:picture filter}
\end{figure}

\subsection{Case Study~(RQ5)}

To answer RQ5, we conduct two case studies on a nature graph and an image to validate the capability of \M~of \texttt{Katz}, \texttt{Logarithm}, and \texttt{Scale-1} graph filters.

i) We tested these three filters on Cora and counted their eigenvalues and frequency of $\hat {A} $ in Figure \ref{fig:cora_eleg}. We found that there are still many nodes in the high-frequency signal range $[1.0, 1.99]$ (frequency 1.99 was too low to display) of Raw, but after filtering, whether it is \texttt{Katz}, \texttt{Logarithm}, or \texttt{Scale-1}, the distribution of eigenvalues shows a long tail distribution. The frequency of high eigenvalue nodes (corresponding to high-frequency signals) is significantly reduced, indicating that the high-frequency signals have been filtered out. Moreover, it is worth noting that different filters have different ranges of eigenvalues after filtering. For example, when \texttt{Katz} is low-pass, its eigenvalue range is [0, 1/($\lambda_{max} \beta_0)]$,  so \texttt{Katz}'s eigenvalues' range is from 0 to $1/(1-(1.99-1)\beta_0)\approx 5$ when $\beta_0=0.8$. In addition, compared to \texttt{Logarithm} and \texttt{Scale-1}, \texttt{Katz}'s eigenvalue distribution curve is smoother when $\beta_0=0.8 $, which proves that \texttt{Katz}'s effect is more effective at this blend factor.

ii) Specifically, given an image with grey values from $0$ to $255$, we first construct a graph in which nodes and edges are the pixels and the links between the nearest 8 neighboring pixels respectively. Then we apply these three filters to the image-based graph. Finally, we reconstruct the image through the filtered graph structure and Figure~\ref{fig:picture filter} depicts filtered images with different $\beta_0$. According to the visualization, we can derive the following observations: First, we observe that for different graph filters under different $\beta_0$, the type and degree are various too. To name some, for \texttt{Katz} and \texttt{Scale-1}, their low-pass effect intensifies with the increase in $\beta_0$. However, as for \texttt{Logarithm}, it is a high-pass filter when $\beta_0 \in (0,0.5)$ while it becomes a low-pass filter when  $\beta_0 \in (0.5, 1)$,  which verifies our discussion in Section~\ref{Discussion of Filter Type}. Furthermore, when comparing \texttt{Scale-1} to \texttt{Katz}, we find that the filter effect of \texttt{Scale-1} experiences a rapid decline as $\beta_0$ decreases. This decline can be attributed to \texttt{Scale-1} imposing a more substantial penalty on distant neighbors with small values of $\beta_0$.

\section{Conclusion}
This paper focuses on the design of power series-enhanced GNNs to address the challenges of long-range dependencies and sparse graphs. To ensure the efficiency of our GPFN, a graph filter using convergent power series from the spectral domain is introduced in this paper. The effectiveness of our GPFN is verified by theoretical analysis from both spectral and spatial perspectives and experimental results, demonstrating its superiority over state-of-the-art graph learning techniques on benchmark datasets. Future directions for investigation include exploring diverse filters such as the mid-pass filter and integrating the diffusion model to further bring the explanation into our GPFN. Moreover, we would like to deeper explore and analysis GPFN on heterogeneous graphs.

\bibliographystyle{named}
\bibliography{ijcai24}

\appendix
\newpage

\section{Prove for Eigenvalue}
In this section, we prove that the maximum of $\widetilde{L}_{sym}$ eigenvalue is $2$ iff the graph is bipartite as bellow: 
\begin{equation}
 \begin{aligned} 
& R(\widetilde{L}_{sym}, u_i) 
        =\frac{u_i^T \widetilde{L}_{sym} u_i}{u_i^T u_i}
        =  \frac{u_i^T (I-\widetilde{D}^{-1/2}\widetilde{A}\widetilde{D}^{-1/2}) u_i}{u_i^T u_i}\notag
        \\
        & = \frac{u_i^T \widetilde{D}^{-1/2}(\widetilde{D} - \widetilde{A})\widetilde{D}^{-1/2} u_i}{u_i^T u_i}= \frac{q_i^T (\widetilde{D} - \widetilde{A}) q_i}{(\widetilde{D}^{1/2}q_i)^T (\widetilde{D}^{1/2}q_i)}\notag  \\
        &  
        = \frac{1}{2} \frac{\sum_{i, j=1}^N \widetilde{A}_{ij}  (q_i - q_j)^2}{\sum_{i=1}^N q_i^2 \widetilde{D}_{ii}}.
 \end{aligned}
    \label{eq:Largest Eigenvale of L_sym 2}
\end{equation}

Then, we apply Cauchy-Schwartz inequality to this equation, and we have:
\begin{equation}
 \begin{aligned} 
        & R(\widetilde{L}_{sym}, u_i)   \leq 
        \frac{\sum_{i,j=1}^N \widetilde{A}_{ij}(q_i^2+q_j^2)}{\sum_{i=1}^N q_i^2 \widetilde{D}_{ii}}
        = \frac{2\sum_{i,j=1}^N \widetilde{A}_{ij} q_i^2}{\sum_{i=1}^N q_i^2 \widetilde{D}_{ii}} \\ &
        = \frac{2\sum_{i=1}^N 
        u_i ^2 / \widetilde{D}_{ii} \sum_{j=1}^N \widetilde{A}_{ij}
        }{\sum_{i=1}^N (q_i \widetilde{D}^{1/2})^2}
        = 2 \frac{ \sum_{i=1}^N \frac{u_i^2}{\widetilde{D}_{ii}}\widetilde{D}_{ii} 
        }{\sum_{i=1}^N u_i^2} \\ &
        = 2 \frac{\sum_{i=1}^N u_i^2
        }{\sum_{i=1}^N u_i^2}
        = 2.
     \end{aligned}
    \label{eq:Largest Eigenvale of L_sym 1}
\end{equation}

Thus, when the graph $\mathcal{G}$ is bipartite, the equality holds as $\sum_{i,j=1}^N \widetilde{A}_{ij} (q_i -q_j)^2 = \sum_{i,j=1}^N \widetilde{A}_{ij} (q_i^2 + q_j^2)$. 

\section{Prove for Mitigating Over-Smoothing}

\subsubsection{Compared to Monomial Graph Filter}
In this section, we compare our filter with the traditional graph filter GCN. Recall that message-passing framework can be described as $H^{(K+1)} = \sigma(\hat A H^{(K)}W^{(K)})$. For better comparison, we omit the activation function $\sigma(\cdot)$ and the learnable weight matrix $W^{(K)}$~\cite{SGC}, we have $H^{(K)} = \hat A^{(K)} X$. As for. GCN we have $H^{(K)} = \hat A^{(K)} X = U^T (\Lambda)^K U X$. Here we taking \texttt{Scale-1} for example, we have $H^{(K)} = \hat A^{(K)} X = U^T F_{\gamma}(\Lambda)^{K} U X$. 

After applying $K$ ($K$ is large enough) GNN layers, the over-smoothing phenomenon occurs. That's to say, the graph embeddings are gradually approaching consensus. Here we assume that the difference between graph embeddings $H^{(K)}$ approach 0, and the embedding of the entire graph can be represented by a full one-vector multiplied by the embedding bound $\mathcal{B}$: $H^{(K)}=\mathcal{B}+o_K(1)$, where $\mathcal{B}$ is a low-rank matrix.

As a consequence, for GCN, we have $H^{(K)}_{:j(\text{GCN})}=\mathbf{1 [b_{\text{GCN}}]} + o_K(1)$ and for \texttt{Scale-1} we have $H^{(K)}_{:j({\text{Scale-1}})}=\mathbf{1 [b_{\text{Scale-1}}]} + o_K(1)$ where $j$ denotes the $j$-th node, $\mathbf{1 [b_{\text{Scale-1}}]}, \mathbf{1 [b_{\text{GCN}}]}$ is the boundary when $K\to +\infty$ for \texttt{Scale-1} and GCN respectively. Next, we compare the convergence rates when different graph filters tend to approach bound with the Euclidean norm:
\begin{equation}
\begin{split}
        & \lim_{K \to +\infty} \frac{H^{(K)}_{:j({\text{Scale-1}})}- \mathbf{1 [b_{\text{Scale-1}}]}}{{H^{(K)}_{:j(\text{GCN})}-\mathbf{1 [b_{\text{GCN}}]}}} 
        \\ &= \lim_{K \to +\infty} \frac{f^{(K)}_\gamma(\hat{A}_{:j(\text{{Scale-1}})}) X- \mathbf{1 [b_{\text{{Scale-1}}}}]}{{\hat{A}^{(K)}_{:j(\text{GCN})}X-\mathbf{1 [b_{\text{GCN}}]}}}
        \\ & =\sum_{i=1}^{N} {\lim_{K \to +\infty}} \|{\frac{f^{(K)}_\gamma(\lambda_{i(\text{{Scale-1}})}) [u_i^T u_i x_i]- b_{i({\text{Scale-1}})}}{\lambda^K_{i({\text{GCN}})} [u_i^T u_i x_i]- b_{i({\text{GCN}})}}} \|_2
        \\ & =\sqrt{\sum_{i=1}^{N} {\lim_{K \to +\infty}}{\frac{f^{(K)}_\gamma(\lambda_{i({\text{Scale-1}})}) [u_i^T u_i x_i]- b_{i({\text{Scale-1}})}}{\lambda^K_{i(\text{GCN})} [u_i^T u_i x_i]- b_{i(\text{GCN})}}} }.
\end{split}\label{eq:fake eq 1}
\end{equation}

Since both the numerator and the denominator tend to $0$ as $K\to +\infty$, and $m_i$ is a variable independent of $K$, we use L'Hôpital's rule:
\begin{equation}
\begin{split}
        & \lim_{K \to +\infty} \frac{H^{(K)}_{:j(\text{Scale-1})}- \mathbf{1 [m_{\text{Scale-1}}]}}{{H^{(K)}_{:j(\text{GCN})}-\mathbf{1 [m_{\text{GCN}}]}}} 
        \\ & =\sqrt{\sum_{i=1}^{N} {\lim_{K \to +\infty}}{\frac{f^{(K)}_\gamma(\lambda_{i(\text{Scale-1})}) \ln(f_\gamma(\lambda_{i(\text{Scale-1})}))[u_i^T u_i x_i]}{\lambda^K_{i(\text{GCN})} \ln(\lambda_{i(\text{GCN})}) [u_i^T u_i x_i]}} }
\end{split}\label{eq:fake eq1 2}
\end{equation}
As we stated before, $f_\gamma(\lambda_{i(\text{Scale-1})}) \geq 0$ and has a upper boundary $\mathbf{M} \in \mathbb{R^+}$ because $f_\gamma(\lambda)$ is convergence. Therefore, the value of this limit primarily depends on the limit of the fraction $\frac{f^{(K)}_\gamma(\lambda_{i(\text{Scale-1})})}{\lambda^K_{i(\text{GCN})}}$:
\begin{equation}
\begin{split}
        & \lim_{K \to +\infty} \frac{f^{(K)}_\gamma(\lambda_{i(\text{Scale-1})})}{\lambda^K_{i(\text{GCN})}} \\ 
        & = \lim_{K \to +\infty} (\frac{1}{\lambda_{i(\text{GCN})} (1-\beta_0 \lambda_{i(\text{Scale-1})})})^K
\end{split}\label{eq:fake eq 3}
\end{equation}
As the eigenvalue of aggregation matrix $\lambda_i \in [0,1]$, it's obviously $\lim_{K \to +\infty} \frac{f^{(K)}_\gamma(\lambda_{i({Scale-1})})}{\lambda^K_{i({GCN})}} \to+\infty$. Hence we have $ o_K(1)= H^{(K)}_{({Scale-1})}- \mathbf{1}\mathcal{B}$ is indeed a low order infinitesimal to graph filter as GCN. 

Therefore, compared with GCN, \texttt{GPFN} is a lower order infinitesimal of GCN, and its convergence speed is slower when over-smoothing occurs, which also proves that \texttt{GPFN} can alleviate overfitting.

\subsubsection{Compared to Polynomial Graph Filter}
Following GPR-GNN, we shrink \texttt{GPFN} to Polynomial Graph Filter by replacing $+\infty$ to $K$ and denote it as \texttt{GPFN-}. Therefore, our aim is to explore whether the part of $ k\to+\infty$ will play a role in the mitigation of over-smoothing.

As a consequence, for \texttt{GPFN-}, we have $H^{(K)}_{:j({\text{GPFN-}})}=\mathbf{1 [b_{\text{GPFN-}}]} + o_K(1)$ where $j$ denotes the $j$-th node, $\mathbf{1 [b_{\text{GPFN-}}]}$ is the boundary when $K\to +\infty$ for \texttt{GPFN-}. Same as before, we use \texttt{Scale-1} for comparison and denote \texttt{GPFN-} as \texttt{Scale-1-}. Next, we compare the convergence rates when different graph filters tend to approach bound with the Euclidean norm.

Since both the numerator and the denominator tend to $0$ as $K\to +\infty$, and $m_i$ is a variable independent of $K$, we use L'Hôpital's rule:
\begin{equation}
\begin{split}
        & \lim_{K \to +\infty} \frac{H^{(K)}_{:j(\text{Scale-1-})}- \mathbf{1 [m_{\text{Scale-1-}}]}}{H^{(K)}_{:j(\text{Scale-1})}- \mathbf{1 [m_{\text{Scale-1}}]}} 
        \\ & =\sqrt{\sum_{i=1}^{N} {\lim_{K \to +\infty}}{\frac{f^{(K)}_\gamma(\lambda_{i(\text{Scale-1-})}) \ln(f_\gamma(\lambda_{i(\text{Scale-1-})}))[u_i^T u_i x_i]}{f^{(K)}_\gamma(\lambda_{i(\text{Scale-1})}) \ln(f_\gamma(\lambda_{i(\text{Scale-1})}))[u_i^T u_i x_i]}} }
        \\ & = 
        \sqrt{\sum_{i=1}^{N} {\lim_{K \to +\infty}} \Bigl(1 - (\frac{\sum_{n=K}^{+\infty}\gamma_n \hat{A}^n X}{\sum_{n=0}^{+\infty}\gamma_n \hat{A}^n X} )^K \Bigl )}.
\end{split}\label{eq:fake eq 2}
\end{equation}
Since $\frac{\sum_{n=K}^{+\infty}\gamma_n \hat{A}^n X}{\sum_{n=0}^{+\infty}\gamma_n \hat{A}^n X} >0$, this equation will approach 0. As a consequence: 
\begin{equation}
\begin{split}
        & \lim_{K \to +\infty} \frac{f^{(K)}_\gamma(\lambda_{i(\text{Scale-1-})})}{f^{(K)}_\gamma(\lambda_{i(\text{Scale-1})})} \\ 
        & = \sqrt{\sum_{i=1}^{N} {\lim_{K \to +\infty}} \Bigl(1 - (\frac{\sum_{n=K}^{+\infty}\gamma_n \hat{A}^n X}{\sum_{n=0}^{+\infty}\gamma_n \hat{A}^n X} )^K \Bigl )} \\
        & = 0.
\end{split}\label{eq:fake eq 333}
\end{equation}
Hence \texttt{GPFN-} is indeed a high order infinitesimal to graph filter as \texttt{GPFN}. 

Therefore, compared with the polynomial graph filter, \texttt{GPFN} is a lower order infinitesimal of \texttt{GPFN-}, and its convergence speed is slower when over-smoothing occurs, which also proves that \texttt{GPFN} can alleviate overfitting.

\section{Baselines}

\begin{table*}[ht]
\centering
\caption{Baseline Code URLs of Github Repository}
\begin{tabular}{l l}
\hline
\textbf{Baseline} & \textbf{Code Repo URL} \\
\hline
LP & \url{https://github.com/sahipchic/VK-LabelPropogation} \\
GCN & \url{https://github.com/tkipf/gcn} \\
GAT & \url{https://github.com/PetarV-/GAT} \\
GIN & \url{https://github.com/weihua916/powerful-gnns} \\
AGE & \url{https://github.com/thunlp/AGE} \\
SGC & \url{https://github.com/Tiiiger/SGC} \\
ChebGCN & \url{https://github.com/mdeff/cnn_graph} \\
GRP-GNN & \url{https://github.com/jianhao2016/GPRGNN} \\
APPNP & \url{https://github.com/benedekrozemberczki/APPNP} \\
BernNet & \url{https://github.com/ivam-he/BernNet} \\
GCNII & \url{https://github.com/chennnM/GCNII} \\
ADC & \url{https://github.com/abcbdf/ADC} \\
DGC & \url{https://github.com/yifeiwang77/DGC} \\
GRAND & \url{https://github.com/THUDM/GRAND} \\
D$^2$PT & \url{https://github.com/yixinliu233/D2PT} \\
HiGNN & \url{https://github.com/Yiminghh/HiGCN} \\
\texttt{GPFN} & \url{https://github.com/GPFN-Anonymous/GPFN} \\
\hline
\end{tabular}
\label{tab:repo}
\end{table*}

\begin{itemize}[leftmargin=*]
\item \textbf{MLP}~\cite{mlp}: MLP simply utilizes the multi-layer perception to perform node classification.
\item \textbf{LP}~\cite{zhuѓ2002learning}: The method predicts the node class by propagating the known labels in the graph, which does not involve processing node attributes.
\item \textbf{GCN}~\cite{GCN}: GCN is a scalable approach for semi-supervised learning on graph-structured data.
\item  \textbf{GAT}~\cite{GAT}: GAT is a spatial domain method, which aggregates information through the attention-learned edge weights.
\item \textbf{GIN}~\cite{gin}: GIN utilizes a multi-layer perceptron to sum the results of GNN and learns a parameter to control residual connection.
\item \textbf{AGE}~\cite{AGE}: AGE applies a designed Laplacian smoothing filter to better alleviate the high-frequency noises in the node attributes.
\item \textbf{SGC}~\cite{SGC}: SGC is a fixed low-pass filter followed by a linear classifier that reduces the excess complexity by removing nonlinearities and weight matrices between consecutive layers. We combine GCN and GAT with SGC to derive \textbf{GCN-SGC} and \textbf{GAT-SGC} for comparison.
\item \textbf{ChebGCN}~\cite{chebnet}: ChebGCN is a graph convolutional network that leverages Chebyshev polynomials for efficient graph filtering and representation learning.
\item \textbf{GPR-GNN}~\cite{GPRGNN}: GPR-GNN learns the weights of representations after information propagation with different steps and performs weighted sum on representations.
\item \textbf{APPNP}~\cite{APPNP}: APPNP approximates topic-sensitive PageRank via a random walk to perform information propagation.
\item \textbf{Res}~\cite{ResGCN}: Res avoids excessive smoothness through residual connection. Like SGC~\cite{SGC}, we also remove nonlinear functions and learnable weight to simplify the Res framework. Besides, we also incorporate GCN and GAT into the Res framework as \textbf{Res-GCN} and \textbf{Res-GAT} for comparison.
\item \textbf{BernNet}~\cite{BernNET}: BernNet uses K-order Bernstein polynomials to approximate graph spectral filters and then performs information aggregation by designing polynomial coefficients.
\item 
\textbf{GCNII}~\cite{chen2020simple}: GCNII is an extension of
the vanilla GCN model with two simple yet effective techniques-- Initial residual and Identity mapping. 
\item 
\textbf{ADC}~\cite{NEURIPS2021_c42af2fa}: ADC learns a dedicated propagation neighborhood for each GNN layer
and each feature channel, making the GNN architecture fully coupled with graph
structures—the unique property that differs GNNs from traditional neural networks.
\item  
\textbf{DGC}~\cite{wang2021dissecting}: DGC decouples the
terminal time and the feature propagation steps, making it more flexible and capable of exploiting a very large number of feature propagation steps.

\item \textbf{GRAND}~\cite{2022GRAND}: A generalized forward push (GFPush) algorithm in GRAND+ to pre-compute a general propagation matrix to perform GNN.
\item \textbf{D$^2$PT}~\cite{10.1145/3580305.3599410}: D$^2$PT performs the dual-channel diffusion message passing with the contrastive-enhanced global graph information on the sparse graph.
\item \textbf{HiGNN}~\cite{HiGCN}: HiGNN proposes a higher-order graph convolutional network grounded in Flower-Petals Laplacians to discern complex features across different topological scales.
\item \textbf{HiD-GCN}~\cite{li2023generalized}: A high-order neighbor-aware graph diffusion network.
\end{itemize}

\section{Hyper-parameter Settings of Baselines}
For GPR-GNN, HiGNN and HiD-GCN, we use the officially released code and other baseline models are based on Pytorch Geometric implementation ~\cite{fey2019PyTorchGeometric}. Table ~\ref{tab:repo} shows the code we used.

The parameters of baselines are also optimized using the Adam with $L_2$ regularization. We set the learning rate at 0.002 with a weight decay of 0.005. Besides, we employ the early-stopping strategy with patience
equal to 20 to avoid over-fitting. 

For MLP, we use 2 layers of a fully connected network with 32 hidden units. For GCN, we use 2 GCN layers with 16 hidden units. For GAT, the first layer has 8 attention heads and each head has 8 hidden units, and the second layer has 1 attention head and 16 hidden units. For GIN, we use two layers with 16 hidden units. For ChebGCN, we use 2 propagation steps with 32 hidden units in each layer. For APPNP, we use a 2-layer MLP with 64 hidden units and set the propagation step K to 10. For GPR-GNN, we use a 2-layer MLP with 64 hidden units and set the propagation steps K to 10, and use PPR initialization. For BernNet, we use a 2-layer MLP with 64 hidden units and set the propagation step K to 10. For SGC, we set layers at K=3.


\section{Training Time Comparison}
\begin{table}[htb]{
\caption{Training time comparison between GPFN and baselines on Cora, Citeseer and AmaComp datasets with masking ratios (MR) equals 0.}
\aboverulesep=0ex
\belowrulesep=0ex
\centering
\label{fig: time}
\begin{tabular}{c|cccc}
\toprule
\multicolumn{5}{c}{Avg. Training Time (ms) / Epoch}    \\
\midrule
\multicolumn{2}{c|}{Dataset}   &  \quad Cora            & \quad Citeseer          & \quad AmaComp        \\ \midrule
\multirow{9}{*}{\rotatebox{90}{Baselines}} & \multicolumn{1}{c|}{GCN} & \quad 25.64        &\quad 25.93         &\quad 24.41         \\
& \multicolumn{1}{c|}{GCN-SGC} & \quad 25.79        &\quad 29.33         &\quad 48.57         \\
& \multicolumn{1}{c|}{APPNP} & \quad 35.14        &\quad 32.40         &\quad 41.69         \\
& \multicolumn{1}{c|}{GRAND} & \quad 32.18        &\quad 34.89        &\quad 53.04         \\
& \multicolumn{1}{c|}{GPRGNN} & \quad 30.15        &\quad 27.75         &\quad 24.86         \\
& \multicolumn{1}{c|}{BernNet} & \quad 25.79        &\quad 29.33         &\quad 48.57         \\
& \multicolumn{1}{c|}{HiGCN} & \quad 56.89        &\quad 59.44         &\quad 72.13         \\
& \multicolumn{1}{c|}{HiD-GCN} & \quad 76.53        &\quad 86.22         &\quad 92.09         \\
& \multicolumn{1}{c|}{GCNII} & \quad 91.62        &\quad 91.63         &\quad 86.95         \\  \midrule
\multirow{3}{*}{\rotatebox{90}{GPFN}} & \multicolumn{1}{c|}{GCN-\texttt{Katz}} & \quad 27.71        &\quad 29.75         &\quad 27.44 \\      
& \multicolumn{1}{c|}{GCN-\texttt{S2}} & \quad 29.89        &\quad 26.64         &\quad 32.52 \\      
& \multicolumn{1}{c|}{GCN-\texttt{S3}} & \quad 30.03        &\quad 26.87         & \quad 35.92 \\      
    \bottomrule
\end{tabular}}
\label{table:time cost}
\end{table}
We compare our GPFN with some SOTA methods that address over-smoothing problems, and the runtime per epoch is shown in Table \ref{fig: time}. It is worth noting that despite our methods making the graph denser, the computational time remains relatively unchanged compared to other methods. GPFN keeps a better balance between time and performance.

\section{Complexity Analysis}

Due to matrix eigenvalue decomposition and matrix inverse operations involved in our computation, the time complexity of the entire process is $O(N^3)$, this is the same with GPRGNN, HiGCN, HiD-GCN and GCNII. However, our GPFN takes into account long-range dependencies, achieving excellent performance without the need for deeper GNN layers, significantly reducing the model's parameter. In contrast, GCNII requires 64 layers to achieve the best performance on Cora.

\section{Motivation Explanation}
The main motivation of our paper lies in addressing two critical challenges faced by existing GNNs: \textbf{handling long-range dependencies} and \textbf{mitigating the adverse impacts of graph sparsity}. 
    Motivated by the power function, we provide a solution that effectively captures long-range dependencies while leveraging the sparse nature of real-world graphs to enhance representation learning.  The proposed GPFN framework is analyzed from both spectral and spatial domains. In the spectral domain, it serves as a flexible graph filter framework capable of accommodating different filter types. In the spatial domain, it acts as an infinite information aggregator, leveraging power series filters to aggregate neighborhood information across an infinite number of hops. This enables the enlargement of the graph's receptive field, thereby addressing the challenge of capturing long-range dependencies effectively.

\end{document}